# Author Accepted Manuscript

## Surrogate-based Optimization via Clustering for Box-Constrained Problems

**Maaz Ahmad and Iftekhar A. Karimi**

**Department of Chemical & Biomolecular Engineering**

**National University of Singapore**

**4 Engineering Drive 4, Singapore 117585**

---

### Publication Information

This document is the author-accepted manuscript of the following peer-reviewed journal article:

Ahmad, M., and Karimi, I. A. (2026). Surrogate-based optimization via clustering for box-constrained problems. *Computers & Chemical Engineering*, 207, 109559.

**Final published article:** https://doi.org/10.1016/j.compchemeng.2026.109559

# Surrogate-based Optimization via Clustering for Box-Constrained Problems

Maaz Ahmad and Iftekhar A Karimi[1]

*Department of Chemical & Biomolecular Engineering*

*National University of Singapore*

*4 Engineering Drive 4, Singapore 117585*

**Abstract**

Global optimization of large-scale, complex systems such as multi-physics black-box simulations and real-world industrial systems is important but challenging. This work presents a novel Surrogate-Based Optimization framework based on Clustering (SBOC) for global optimization of such systems, which can be used with any surrogate modeling technique. At each iteration, it uses a single surrogate model for the entire domain, employs k-means clustering to identify unexplored domain, and exploits a local region around the surrogate's optimum to potentially add three new sample points in the domain. SBOC has been tested against sixteen promising benchmarking algorithms using 52 analytical test functions of varying input dimensionalities and shape profiles. It successfully identified a global minimum for most test functions with substantially lower computational effort than other algorithms. It worked especially well on test functions with four or more input variables. It was also among the top six algorithms in approaching a global minimum closely. Overall, SBOC is a robust, reliable, and efficient algorithm for global optimization of box-constrained systems.

## 1. Introduction

The rapid evolution of computing processors, advances in artificial intelligence, machine learning (ML), internet of things technology, and the desire to remain competitive in the ever-evolving industrial landscape have been driving Industry 4.0. Digital twins or virtual models are at the heart of this industrial transformation. By replicating real-world systems or processes, digital twins enable decision makers to systematically understand, analyze, and assess the impact of various decisions in a safe and inexpensive manner. They are also essential for optimizing real-world industrial systems to achieve desired goals such as efficiency, safety, quality, cost, energy, emissions, etc.

A common practice in optimization has been to use first-principles simulation models as digital twins. However, most commercial simulators are computational black boxes with no access to source codes and no analytical closed-form expressions for optimization objectives or constraints. Numerical differentiation remains the only option for derivative-based optimization (DBO) algorithms, which is expensive, prone to truncation and floating-point errors (Nocedal and Wright, 2009), and unreliable for noisy systems. In other words, derivative-free optimization (DFO) algorithms such as direct search (Kolda et al., 2003) and metaheuristics (Abdel-Basset et al., 2018) become necessary for optimizing black-box systems. These typically require many function evaluations, hence are expensive.

The above discussion suggests that reducing function evaluations is the key to efficient optimization of systems with compute-intensive function evaluations. Hence, surrogate-based optimization (SBO) algorithms (Boukouvala et al., 2016; Forrester and Keane, 2009; Rios and Sahinidis, 2013) offer an attractive option. They substitute objective and/or constraints by inexpensive and accurate data-driven surrogate models during their searches. Their closed-form expressions and inexpensive evaluations allow one to use both DFO and DBO algorithms. Several studies (Boukouvala and Floudas, 2017; Garud et al., 2019; Gutmann, 2001; Müller and Shoemaker, 2014; Regis and Shoemaker, 2005, 2007a; Srinivas and Karimi, 2024) have reported superior performance of SBO algorithms in accurately locating a global optimum at lower computational expense.

[1] Corresponding author: Email cheiak@nus.edu.sg

Recent advances in ML and active learning (Ren et al., 2021) have prompted several promising iterative SBO algorithms. Depending on surrogate scope and search strategy, they can be classified as either local or global methods. Local SBO algorithms typically approximate the objective function within a small region (trust-region) by a simple surrogate model (e.g. linear or quadratic) at each iteration. Using its minimizer, they determine a new iterate and trust region. They repeat this until the trust region shrinks below a user-specified threshold. Local SBO algorithms are highly sensitive to the choices of start point and trust region size. Surrogate Management Framework (Booker et al., 1999), Unconstrained Optimization by Quadratic Approximation (UOBYQA) (Powell, 2002) and Bound Constrained Optimization by Quadratic Approximation (BOBYQA) (Powell, 2009) are examples of such algorithms.

In contrast, global SBO algorithms build one or more surrogates for the objective or constraints over the entire domain using an initial set of well-spaced sample points. The sample point with the best objective value is identified and iteratively improved by adding new points in promising regions to update the surrogate model. The point addition strategy constitutes a key step in these algorithms. While it is important to search locally around the current best point (local exploitation) to accelerate convergence to the optimum, it is equally important to search over the entire domain (global exploration) to avoid local traps. By capturing the trend and topography of the objective, surrogate models guide the placement of new points to explore and/or exploit the domain better. Thus, global SBO algorithms are better suited for locating the global optima than local algorithms.

Efficient Global Optimization (EGO) by (Jones et al., 1998) and Gutmann's algorithm (Gutmann, 2001) were the earliest global SBO algorithms that laid the foundation for subsequent developments. EGO exploits the ability of Kriging (KRG) technique to quantify uncertainty in model predictions in terms of an expected improvement (EI) function. EI computes the improvement expected in the global optimum estimate from a new point at a specific location. EGO maximizes EI at each iteration to efficiently explore under-sampled regions and exploit regions with promising objective values. Several later works addressed the limitations of EGO and extended its capabilities. Forrester and Jones, 2008 observed that EI may place points sub-optimally during initial iterations, as initial KRG surrogates are less accurate. Hence, they modified EI to strongly favor exploration during initial iterations. Boukouvala and Ierapetritou, 2014; Huang et al., 2006; Quan et al., 2013 adapted EGO for noisy systems. Assuming homoscedastic noise variance throughout the domain, Huang et al., 2006 proposed an augmented EI (AEI) function that accounts for system noise by allowing replicates to reduce prediction uncertainty. Quan et al., 2013 proposed adding a new point based on EI and multiple replicates based on the optimal budget allocation strategy of Chen et al., 2000 at each iteration to optimize systems with heterogenous noise variance. Boukouvala and Ierapetritou, 2014 extended Huang et al.'s, 2006 algorithm to stochastic systems with linear and nonlinear constraints. Zhan et al., 2017 defined a pseudo-EI (PEI) function by combining the original EI with an influence function that estimates the expected improvement from incremental point additions. They maximized PEI at each iteration to add multiple points and used parallel computations to accelerate convergence. In a similar vein, Xing et al., 2020 proposed a multi-point sampling strategy by coupling EI with adaptive domain reduction. Non-EI based point addition strategies have also been reported. Villemonteix et al., 2009 proposed an entropy measure. Davis and Ierapetritou, 2008 designed a two-step KRG-dependent SBO algorithm that differed from EGO. Their first step iteratively improved the surrogate accuracy by using the centroids of Delaunay triangulations (Delaunay, 1934) (formed by the existing sample points) at each iteration. Their second step used local quadratic models around the new points to locate local optima.

Gutmann's algorithm used Radial Basis Functions (RBFs) as surrogates. It hypothesizes that the RBF surrogate trained on the existing sample points passes through an estimate of the global minimum at each iteration. The credibility of this hypothesis is strengthened, when the surrogate response surface varies smoothly across the domain. To this end, Gutmann, 2001 quantified the 'bumpiness' of an RBF surrogate and minimized it at each iteration to add new points. However, the algorithm converges slowly, when global minimum is trapped in steep and narrow basins. Regis and Shoemaker, 2006 addressed this limitation by minimizing the bumpiness function locally around the point with the best objective value, rather than over the entire domain. Regis and Shoemaker, 2005 also developed an algorithm (CORS-RBF) that differed from Gutmann's algorithm. This algorithm simply minimizes the RBF surrogate, while ensuring minimum separation between neighboring points to add new points iteratively. Later, Regis and Shoemaker, 2007b proposed a master-worker framework for both their algorithms to sample multiple points at each iteration and use multiprocessing to speed up convergence. Jakobsson et al., 2010 proposed an algorithm to optimize noisy objectives using RBF surrogates. They defined a utility function to efficiently balance exploration versus exploitation for point addition. More recently, Yi et al., 2020 developed an algorithm for optimizing objectives that can be represented via high- and low-fidelity approximates. Their algorithm uses a typical SBO framework based on RBF technique and a weighted metric for iterative sampling. The framework first searches for a global optimum over the low-fidelity approximate until a user-specified number of points is sampled. Then, the surrogate trained is minimized using a gradient-based solver to determine its local minimizers. These minimizers are used as initial points for a second round of SBO over the high-fidelity approximate.

While the above algorithms have demonstrated much promise, their primary limitation is their dependency on specific surrogate modeling techniques such as KRG or RBF. A decision-making process may often demand alternate modeling techniques due to factors such as model interpretability, suitability, simplicity, generalizability, or simply familiarity. Thus, generic SBO algorithms that are agnostic to the choice of modeling techniques are indeed desirable. Some works have addressed this issue. Regis and Shoemaker, 2007a proposed a multi-start local metric stochastic response surface (MLMSRS) algorithm that adds points in a local neighborhood of the best current point using a weighted metric. This metric explores sparsely sampled regions and exploits regions with promising surrogate predictions. The local search size expands, if a better point is identified, otherwise it contracts. Once this size falls below a threshold, the algorithm starts afresh with a new set of initial points. Beyhaghi et al., 2016 developed an algorithm that works with any interpolating modeling technique (KRG, RBF, polynomial spline, …). It constructs Delaunay triangulations of the sample points to explore triangulations with large circumradii. It also exploits regions near the local minimizers of the interpolating model. Garud et al., 2019 also used Delaunay triangulations at each iteration. They balanced exploration and exploitation using a weighted metric to select a triangulation for placing a single new point. They solved a surrogate-based optimization sub-problem for this placement. More recently, Ma et al., 2023 proposed the Branch-and-Model (BAM) algorithm based on dynamic domain partitioning and local surrogate modeling. At each iteration, BAM divides the domain into exhaustive boxes (hyper-rectangles) such that each box comprises exactly one point, selects a few potentially optimal boxes based on DIRECT’s identification scheme, builds a distinct local surrogate over each selected box using the ALAMO (Cozad et al., 2014a) surrogate builder framework, minimizes each local surrogate using the BARON solver (Sahinidis, 1996), and evaluates the objective function at the most promising candidate points determined during surrogate minimization. These steps continue until a user-specified limit on maximum function evaluations is exhausted. With a similar approach, Srinivas and Karimi, 2024 proposed a zone-wise SBO (ZSBO) algorithm that approximated the search domain via multiple local surrogates for distinct regions. However, unlike BAM, ZSBO did not necessarily define boxed regions. Rather, it used a clustering technique to identify these regions and added at most four new points at each iteration. While these generic SBO algorithms were aimed at box-constrained systems, others have targeted constrained (Beykal et al., 2022; Boukouvala and Floudas, 2017; Kleijnen et al., 2010; Regis, 2011) and mixed-integer nonlinear (Karia et al., 2025; Kim and Boukouvala, 2020; Mistry et al., 2021; Muller, 2016; Rashid et al., 2012) optimization problems.

Existing clustering-based SBO algorithms have employed clustering for various purposes. Villanueva et al., 2013 used it to define a local region around each current distinct local optimum. Their SBO framework added one new point in each region at each iteration by minimizing a local surrogate. Boukouvala and Floudas, 2017 defined a spherical region around each distinct local optimum (Boukouvala et al., 2017) and gradually tightened its size to refine optimal solutions in their global optimization framework (ARGONAUT). Dong et al., 2018 used k-means clustering (MacQueen, 1967) to identify sparse regions, when the best solution failed to improve for some successive iterations. They relied on the user to specify the number of clusters and minimum cluster size. However, they placed new points *outside* the clusters using Latin Hypercube Design. Srinivas and Karimi, 2024 applied k-means clustering to construct their multiple local surrogates. Several direct search algorithms (Boender et al., 1982; Csendes et al., 2008; Torn, 1987) have used hierarchical clustering (Mojena, 1977) to identify ‘basins of attraction’ hosting different local optima in order to seek a global optimum.

Despite increased interest, optimizing large-scale, complex, and expensive black-box models (especially those with high-fidelity multi-physics simulation) to global optimality remains a significant challenge and an active area of research. This work focuses on this challenge. We propose a novel global SBO algorithm for optimizing box-constrained systems in a robust, reliable, and computationally efficient manner. Our algorithm leverages clustering to define local regions and iteratively samples multiple new points in the most favorable locations of the domain. We call our proposed algorithm SBOC or Surrogate-Based Optimization via Clustering. Also 'SBO' will henceforth mean global SBO in this paper. Some of the key features of SBOC are as below.

1. It is model agnostic. It works with any modeling technique or surrogate form.

2. It uses a single global surrogate model versus multiple local surrogates (Ma et al., 2023; Srinivas and Karimi, 2024). This choice avoids several limitations associated with using multiple local surrogates such as the need for efficient domain partitioning, defining exhaustive local regions, choosing simple yet accurate surrogates for individual regions, determining surrogate influence over overlapping regions, and limiting computational cost of constructing multiple local surrogates at each iteration.

3. It uses clustering to explore the domain thoroughly and exploit promising regions systematically. The rationale for clustering points in SBOC differs fundamentally from prior algorithms that have employed clustering to identify basins of attraction (Boender et al., 1982; Csendes et al., 2008; Torn, 1987), define regions around known local optima (Boukouvala and Floudas, 2017; Villanueva et al., 2013), escape local optima (Dong et al., 2018), or construct multiple local surrogates for different zones in the domain (Srinivas and Karimi, 2024).

4. It defines multiple local regions in the domain by clustering the sample points at each iteration. This geometric partitioning approach is simpler, faster, and computationally cheaper than the expensive and inefficient (Goodman et al., 2017) Delaunay triangulations (Beyhaghi et al., 2016; Davis and Ierapetritou, 2008; Garud et al., 2019).

5. It determines three new points at each iteration. Each point is targeted exclusively for either exploration or exploitation. This is in contrast to the alternate approach of using a weighted metric for balancing exploration versus exploitation as done by EI-based methods, Garud et al., 2019; Regis and Shoemaker, 2007b, 2007a, 2005.

After defining the problem in Section 2, we detail the key steps of SBOC in Section 3. We illustrate its application in Section 4 using a test function first and then assess its broader effectiveness in Section 5 via an extensive numerical exercise. Section 6 compares SBOC's performance versus existing SBO algorithms. Section 7 concludes our work.

## 2. Problem Statement

We aim to develop a robust, reliable, and computationally efficient SBO algorithm to solve the following problem.

*Given:*

1. $N$ continuous decision variables / inputs, $x = [x_1, x_2, \dots, x_N] \subset \mathbb{R}^N$ and input domain $D = \left\{ x \mid x_n^{(L)} \le x_n \le x_n^{(U)}\ \forall n = 1, 2, \dots, N \right\}$.

2. Computationally expensive function $f(x \in D)$: $\mathbb{R}^N \to \mathbb{R}$ to evaluate the objective at the sample points. For black-box simulators, a callable executable must be provided that returns the objective function value at any specified $x$.

3. Computational budget expressed as the maximum number ($K_{max}$) of $f(x)$ evaluations.

4. User-specified surrogate form $\tilde{S}(x)$ or modeling technique for building surrogate models and a criterion to judge surrogate quality. While our algorithm works with any surrogate modeling technique, it is desirable to choose a flexible surrogate form that can adequately capture the nonlinearities in the objective function landscape. Promising surrogate recommendation and development frameworks such as FBS (Ahmad and Karimi, 2022), LEAPS2 (Ahmad and Karimi, 2021), AI-DARWIN (Chakraborty et al., 2021), SPA (Sun and Braatz, 2021), CRS (Cui et al., 2016), ALAMO (Cozad et al., 2014b), and SUMO (Gorissen et al., 2010) can guide this selection.

*Obtain:*

A global minimizer $\boldsymbol{x}^* \in D$ that solves the following optimization problem.

$$f^* = f(\boldsymbol{x}^*) = \inf_{x \in D} f(\boldsymbol{x}) \tag{1}$$

## 3. Key Steps of SBOC

With no loss of generality, let $D$ be normalized to $[0,1]^N$. SBOC begins by sampling a few points in $D$. It then evaluates $f(\boldsymbol{x})$ at these points to generate an initial input-output data set. At each iteration, the given $\tilde{S}(\boldsymbol{x})$ is trained on the current data set to build a surrogate model $S(\boldsymbol{x}) \approx f(\boldsymbol{x})$. The points in the data set are then clustered via the k-means technique to distinguish the domain's well-sampled regions from under-explored ones. These clusters along with $S(\boldsymbol{x})$ guide SBOC to add new distinct points in $D$ aimed at exploration and exploitation. $f(\boldsymbol{x})$ is evaluated at the newly added points and the data set is updated. This iterative process of surrogate update, cluster formation, and point addition is repeated until computational budget is exhausted.

In the following subsections, we explain each SBOC step in detail.

### 3.1 Obtain Initial Points

SBOC allows users to provide a set of initial sample points directly or to specify a sampling technique and an initial number ($K_0$) of points to be generated. In the absence of user specifications, SBOC samples $K_0 = 5N$ initial points ($\boldsymbol{x}^{(k)}, k = 1, 2, \dots, K_0$) via Sobol sampling (Sobol', 1967). Garud et al., 2017 have shown Sobol sampling to possess excellent space-filling capabilities, which are desirable for a good initial $S(\boldsymbol{x})$ that captures system behavior across $D$. $f(\boldsymbol{x})$ is evaluated at the sampled points to generate the initial data set $\mathcal{T}_1$: $\left[\boldsymbol{x}^{(k)}, y^{(k)} = f\left(\boldsymbol{x}^{(k)}\right)\right], k = 1, 2, \dots, K_1 (= K_0)$. SBOC then attempts to train $\tilde{S}(\boldsymbol{x})$ on $\mathcal{T}_1$. In case the points are insufficient to develop an initial surrogate (i.e. determine all model parameters), SBOC further samples $p$ points by Sobol sampling, where $p$ is the number of free model parameters in $\tilde{S}(\boldsymbol{x})$.

After the above initialization, SBOC begins its iterations ($i = 1, 2, \dots$). Let $\mathcal{T}_i$ denote the starting data set at iteration $i$ and $K_i$ denote the number of its sample points. Let $\hat{\boldsymbol{x}}_i^*$ and $\hat{f}_i^*$ be defined as follows.

$$\hat{f}_i^* = \min_{1 \le k \le K_i} y^{(k)} \tag{2a}$$

$$\hat{\boldsymbol{x}}_i^* = \arg \hat{f}_i^* \tag{2b}$$

The detailed steps at each iteration $i$ are as follows.

### 3.2 Construct Surrogate $S(\boldsymbol{x})$

Train $\tilde{S}(\boldsymbol{x})$ on $\mathcal{T}_i$ to obtain $S_i(\boldsymbol{x})$. This involves determining the best values of $\tilde{S}(\boldsymbol{x})$'s parameters that optimize a user-specified surrogate quality criterion.

### 3.3 Minimize $S(\boldsymbol{x})$ for First Point

Determine the best optimizer $\hat{\boldsymbol{x}}_i$ of $S_i(\boldsymbol{x})$ using any multi-start optimization algorithm with $\boldsymbol{x}^{(k)}$ ($k = 1,2, \dots, K_i$) as the start points.

$$\hat{\boldsymbol{x}}_i = \arg \inf_{\boldsymbol{x} \in D} S_i(\boldsymbol{x}) \tag{3}$$

Append $[\hat{\boldsymbol{x}}_i, f(\hat{\boldsymbol{x}}_i)]$ to $\mathcal{T}_i$, if $\hat{\boldsymbol{x}}_i^*$ is well-separated from the current sample points in $\mathcal{T}_i$. Euclidean distance can be used as a measure of separation, which must exceed $\varepsilon = 0.0001\sqrt{N}$ from each point. Update $\hat{\boldsymbol{x}}_i^*$ and $\hat{f}_i^*$ in Eq. 2ab.

### 3.4 Distribute $\mathcal{T}_i$'s Points into $C_i^*$ Clusters

Determine $C_i^*$ as follows and use the k-means technique to form $C_i^*$ clusters of the points in $\mathcal{T}_i$. Since the k-means algorithm can accept any number ($C > 1$) of clusters, we optimize $C$ using at each iteration $i$ using the elbow method proposed by Thorndike, 1953. This method exploits the key idea behind the k-means algorithm, which is to classify points into $C$ clusters such that each point is closer to its own cluster's centroid than to any other. This amounts to minimizing the intra-cluster squared distance ($ICSD$) in each cluster, which is the sum of the Euclidean distances of its points from its centroid. As $C$ increases, the total $ICSD$ ($TICSD$) or the sum of $ICSD$s of all $C$ clusters, decreases monotonically. After a certain $C$, its decrease is slow, hence we take $C_i^*$ as the $C$ beyond which $TICSD$ reduces marginally as follows.

$$C_i^* = \min_{1 < C < K_i} C \ \textit{such that} \ \frac{TICSD_C - TICSD_{C+1}}{TICSD_1 - TICSD_2} < 10\% \tag{4}$$

Classify all points in $\mathcal{T}_i$ into $C_i^*$ clusters using the k-means algorithm.

### 3.5 Explore Vacant Sub-Spaces for Second Point

The previous step gives us clusters of points versus local regions or subspaces in $D$. However, we wish to (1) find a point that is in some large inter-cluster subspace devoid of any cluster points, and (2) avoid defining an explicit local region or subspace for each cluster. Our aim is to explore the domain more efficiently than done by expensive geometrical constructions such as convex hulls and Delaunay triangulations that define subspaces or regions explicitly. The following simple heuristic procedure aims to achieve these objectives.

Define $ICD$ or inter-cluster distance between two clusters $u$ $(1 \le u \le C_i^*)$ and $v$ $(v \ne u,\ 1 \le v \le C_i^*)$ as follows.

$$ICD_{uv} = \min_{p,q} \left\| \boldsymbol{x}^{(p)} - \boldsymbol{x}^{(q)} \right\| \tag{5}$$

such that $\boldsymbol{x}^{(p)}$ is a sample point in cluster $c = u$ and $\boldsymbol{x}^{(q)}$ is in cluster $c = v$. Now, for each cluster $c = 1,2,\dots,C_i^*$, identify its nearest cluster $n(c)$ based on the inter-cluster distance between them. This gives us at most $C_i^*$ pairs of nearest neighbors, because $c = n[n(c)]$ may not hold. Now, from these pairs, find the pair with the longest inter-cluster distance. Let $[u^*,\ n(u^*)]$ denote this pair of farthest neighbor clusters. Now, identify the sample points $\boldsymbol{x}^{(p^*)} \in u^*$ and $\boldsymbol{x}^{(q^*)} \in n(u^*)$ that correspond to the inter-cluster distance between $u^*$ and $n(u^*)$. We claim that the points along the line joining these two points likely lies in a large inter-cluster space devoid of cluster points. Hence, select 0.5*$[\boldsymbol{x}^{(p^*)} + \boldsymbol{x}^{(q^*)}]$ as the second point. Evaluate $f(0.5 * [\boldsymbol{x}^{(p^*)} + \boldsymbol{x}^{(q^*)}])$ and update $K_i$, $\mathcal{T}_i$, $\widehat{\boldsymbol{x}}_i^*$, and $\hat{f}_i^*$.

### 3.6 Exploit Around $\widehat{\boldsymbol{x}}_i^*$ for Third Point

The previous step added a point in an under-explored region of $D$. To improve $\widehat{\boldsymbol{x}}_i^*$, we now add a new point in its neighborhood to get a more accurate $S(\boldsymbol{x})$ locally around $\widehat{\boldsymbol{x}}_i^*$. Define a local neighborhood $LN(\widehat{\boldsymbol{x}}_i^*)$ comprising its nearest $ceil[0.2K_i]$ sample points based on Euclidean distance, where the $ceil(z)$ represents the smallest integer $\ge z$. Note that these neighbor points may belong to any clusters or may include the point added in the previous step. Now, add a new point that (1) lies within $LN(\widehat{\boldsymbol{x}}_i^*)$, and (2) is closer to points with lower $f(\boldsymbol{x})$, using the following approach.

Assign a weight $w_l \in [0,1]$ to each sample point $\boldsymbol{x}^{(l)} \in LN(\widehat{\boldsymbol{x}}_i^*)$ using the following, motivated by the characteristic bell-shaped symmetric curve for the Gaussian function $\exp(-(z-\mu)/\sigma^2)$, which peaks at $z = \mu$ and tapers further away depending on $\sigma^2$,

$$w_l = \frac{\exp\left[-\sqrt{f(\boldsymbol{x}^{(l)}) - \hat{f}_i^*}/\eta\right]}{\sum_{l \in LN(\widehat{\boldsymbol{x}}_i^*)} \exp\left[-\sqrt{f(\boldsymbol{x}^{(l)}) - \hat{f}_i^*}/\eta\right]} \tag{6}$$

where, $\eta$ varies cyclically $\{0.5, 1.5, 2.5, 5, 10\}$ through iterations. Eq. 6 assigns a larger weight to points with $f(\boldsymbol{x})$ closer to $\hat{f}_i^*$ to enable a better $S(\boldsymbol{x})$ in that region. Now, select the new point as the linear combination of all $LN(\widehat{\boldsymbol{x}}_i^*)$ points given by $\sum_{l \in LN(\widehat{\boldsymbol{x}}_i^*)} w_l \boldsymbol{x}^{(l)}$, as long as it is not too close (distance > $0.0001\sqrt{N}$) to any existing sample point. Update $K_i$, $\mathcal{T}_i$, $\widehat{\boldsymbol{x}}_i^*$, and $\hat{f}_i^*$.

Varying $\eta$ cyclically across iterations changes the sensitivity of the weights to variations in $f(\boldsymbol{x})$, where a smaller $\eta$ makes $w_l$ more sensitive to finer differences in $f(\boldsymbol{x}^{(l)})$. This strategy of using periodically varying $\eta$ across consecutive iterations ensures that new points are not added repeatedly at the same location, specifically when $\widehat{\boldsymbol{x}}_i^*$ does not improve over successive iterations.

### 3.7 Check for Termination

Terminate if $K_i \ge K_{max}$ and report $\widehat{\boldsymbol{x}}^* = \widehat{\boldsymbol{x}}_i^*$ as an estimated global optimum. Otherwise, $i \leftarrow i + 1$ and re-iterate from Step 3.2. Fig. 1 presents a flowchart for SBOC.

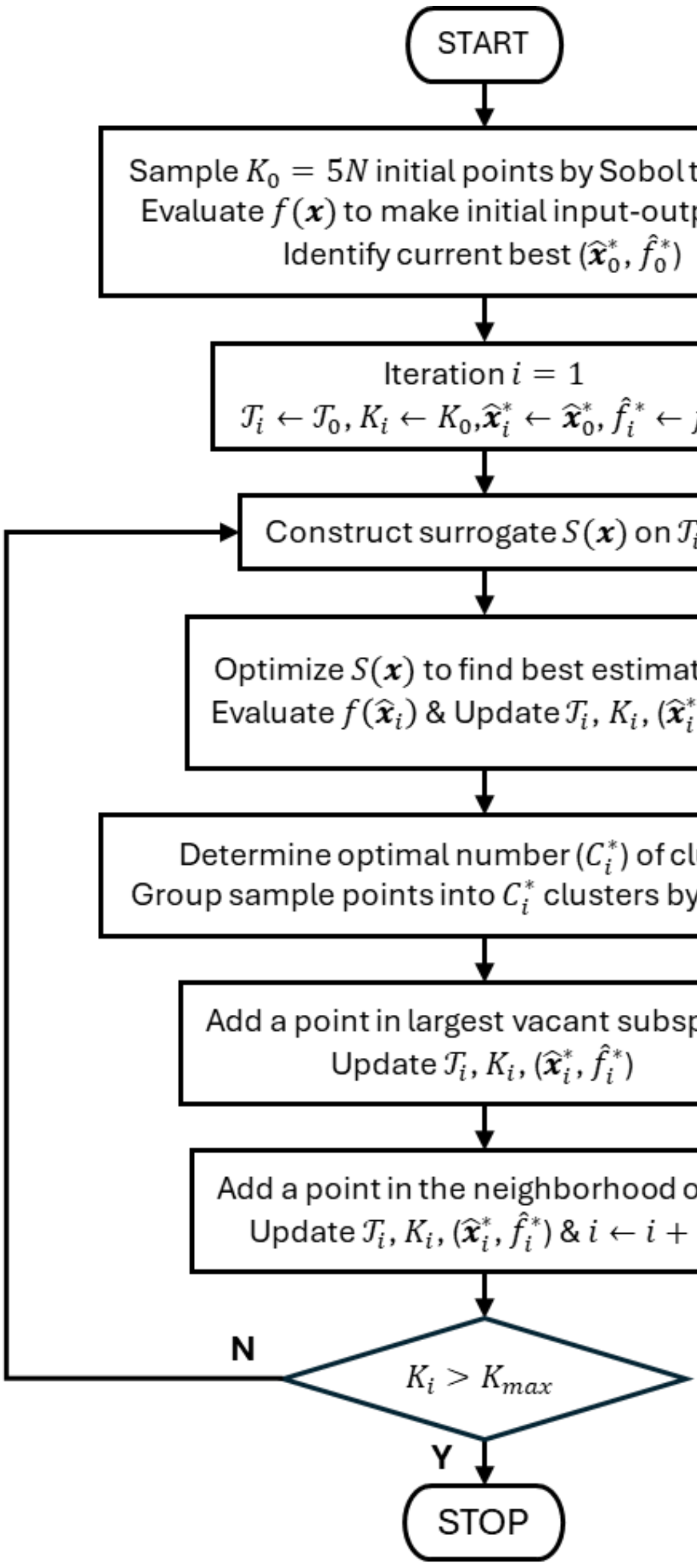

**Fig. 1:** Flowchart for the SBOC algorithm

## 4. SBOC Illustration on a Test Example

Consider minimizing the six-hump camel back function (SHCB) (Surjanovic and Bingham, 2013) below.

$$SHCB: f(\boldsymbol{z}) = \left(4 - 2.1z_1^2 + \frac{z_1^4}{3}\right)z_1^2 + z_1z_2 - (4 - 4z_2^2)z_2^2 \tag{7}$$

SHCB has $N = 2$ input variables with bounds: $-2 \le z_1 \le 2$ and $-1 \le z_2 \le 1$. It has two global minima and four local minima. The global minima occur at $\boldsymbol{z}^* = [0.0898, -0.7126]$ and $[-0.0898, 0.7126]$ with $f^* = -1.0316$. In the normalized domain $D = [0,1]^2$, they occur at $\boldsymbol{x}^* = [0.5225, 0.1437]$ and $[0.4775, 0.8563]$. Fig. 2 shows its surface and contour plots.

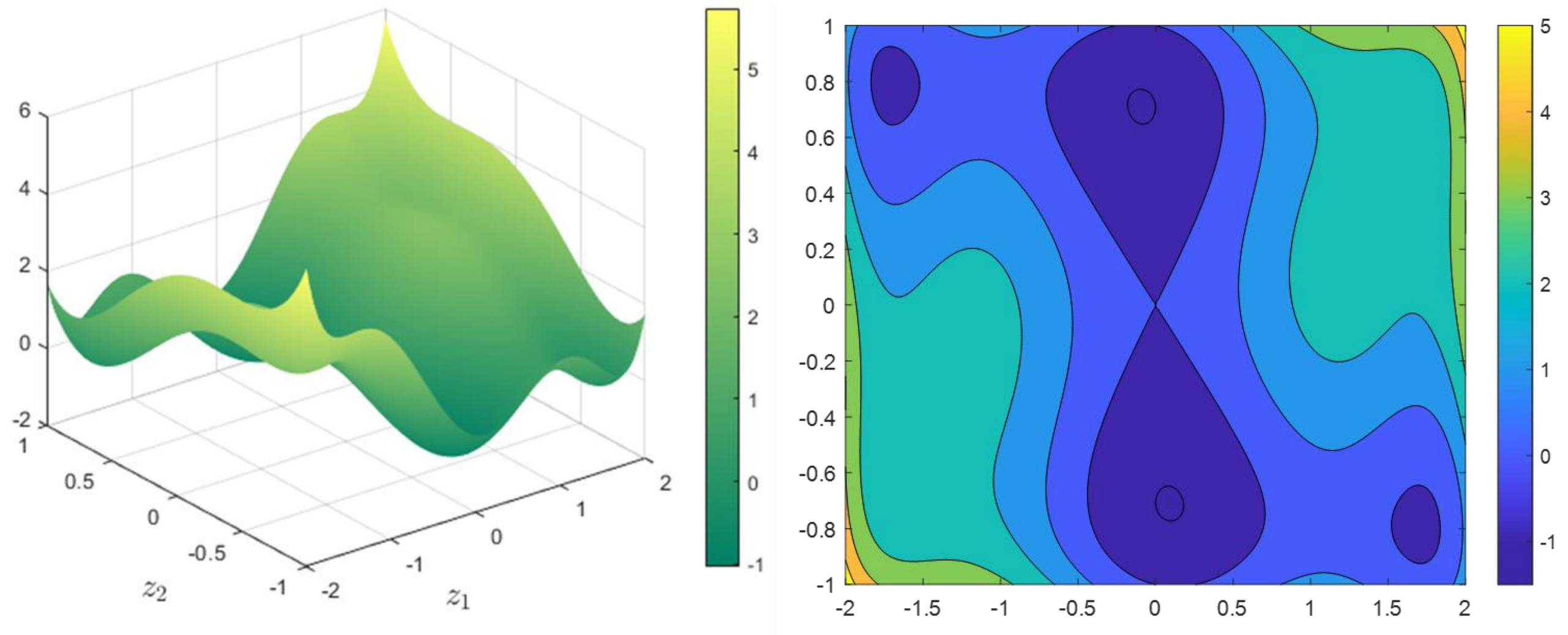

**Fig. 2:** Surface and Contour Plots for Six Hump Camel Back Function

Let $K_{max} = 50$, $K_0 = 5N = 10$, and $\tilde{S}(\boldsymbol{x})$ be an RBF (Radial Basis Function, Hardy, 1971) with a multi-quadratic basis function $\tau_k(\boldsymbol{x})$ and linear tail function $t(\boldsymbol{x})$ as follows.

$$\tilde{S}(\boldsymbol{x}) = \sum_{k=1}^{K} \beta_k \tau_k\left(\left\|\boldsymbol{x} - \boldsymbol{x}^{(k)}\right\|\right) + t(\boldsymbol{x}) \tag{8a}$$

$$\tau_k(\boldsymbol{x}) = \sqrt{\left\|\boldsymbol{x} - \boldsymbol{x}^{(k)}\right\|^2 + \psi^2} \tag{8b}$$

$$t(\boldsymbol{x}) = a_0 + \sum_{n=1}^{N} a_n x_n \tag{8c}$$

where, $\psi$, $\beta_k$ $(k = 1, 2, \dots, K)$, and $a_n$ $(n = 0, 1, \dots, N)$ are parameters. $\psi$ is a hyper-parameter that is prefixed before determining the other parameters for every $S(\boldsymbol{x})$. As suggested by (Acar, 2015), we use 10 equi-spaced $\psi$ values including $\psi = 1/K$ and $\psi = 1$ as end points. For each $\psi$, we train an $S(\boldsymbol{x})$ using $0.8K$ randomly chosen sample points and assess its accuracy on the remaining sample points. The $\psi$ that yields the most accurate model is selected and fixed in $\tilde{S}(\boldsymbol{x})$. Then, the final $S(\boldsymbol{x})$ is trained using all $K$ sample points.

SBOC starts by sampling 10 points (white circles in Fig. 3a) in $D$ via Sobol sampling to get $\mathcal{T}_1$ with $\widehat{\boldsymbol{x}}_0^* = [0.3853, 0.8083]$ (red star in Fig. 3a) and $\hat{f}_0^* = -0.4732$, which is far away from both global minima (green stars in Fig. 3a).

In the first iteration ($i = 1$), we use $\mathcal{T}_1$ to get $S_1(\boldsymbol{x})$ as described before. Minimizing $S_1(\boldsymbol{x})$ using an SQP (Sequential Quadratic Programming) solver (Wilson, 1963) starting from each of 10 points in $\mathcal{T}_1$ gives $\widehat{\boldsymbol{x}}_1 = [0, 1]$ (yellow star in Fig. 3b) as the best surrogate solution. However, the true $f(\widehat{\boldsymbol{x}}_1) = 1.7333$ is not good due to poor surrogate approximation. Nevertheless, $\widehat{\boldsymbol{x}}_1$ is added to $\mathcal{T}_1$ as the first point.

The elbow method suggests making $C_1^* = 4$ clusters from 11 sample points. Fig, 3c shows these cluster points as red ($c = 1$), blue ($c = 2$), green ($c = 3$), and white ($c = 4$). Neighbor cluster pairs turn out to be $(1, 2)$, $(2, 3)$, and $(4, 1)$ with inter-cluster distances of 0.240, 0.236, and 0.373 respectively. Clusters 4 and 1 are farthest apart so the second point should be placed in between them. $[0.6810, 0.1985]$ is the midpoint (yellow star in Fig 3c) on the dashed line (Fig 3c) between $[0.7448, 0.0230]$ in cluster 4 and $[0.6171, 0.3739]$ in cluster 1. It is added as the second point to $\mathcal{T}_1$. $f(\boldsymbol{x}) = 0.2050$ at this point, so the current best point remains unchanged.

Since $K_1 = 12$, $LN(\widehat{\boldsymbol{x}}_1^*)$ should have $ceil[0.2 \times 12] = 3$ points. These are marked black in Fig. 3d. Using $\eta = 0.5$ and computing weights using Eq. 6, we get $[0.4021, 0.8590]$ as the third point using Eq. 7 with $f(\boldsymbol{x}) = -0.7150$. $\mathcal{T}_1$ now has 13 points, $\widehat{\boldsymbol{x}}_1^* = [0.4021, 0.8590]$, and $\hat{f}_1^* = -0.7150$. The red star in Fig. 3e marks this new best point. Iteration 1 has ended.

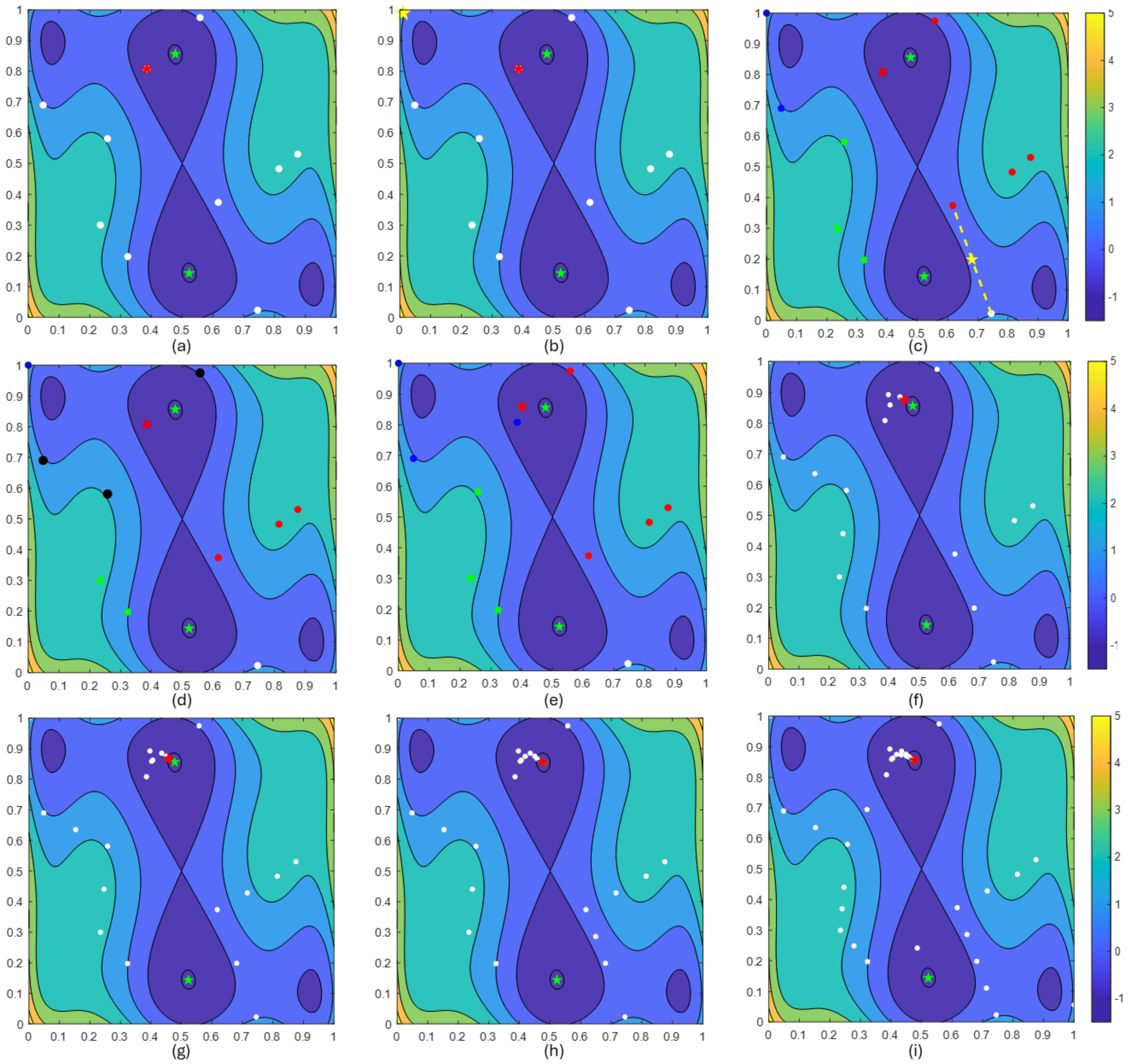

**Fig. 3:** Illustration of key SBOC steps and their progression on Six Hump Camel Back Function. (a) Initial points (white circles), true global minima (green stars), and current minimum (red star). (b) First point (yellow star) added by surrogate minimization. (c) Color-coded cluster points (red, blue, green, white) and Second point (yellow star) added at the midpoint of the points defining inter-cluster distance (dashed line). (d) Neighbor points (black circles) around current minimum. (e) Third point (red star), also the current minimum, added by linear combination of neighbor points. (f-i) Current minimum after 3 (f), 4 (g), 5 (h), and 10 (i) iterations

Repeating the above steps, SBOC continues for 17 iterations, when $K_{17} = 51$ exceeds $K_{max} = 50$. Its best estimate for a global minimum is $\hat{\boldsymbol{x}}^* = [0.4776, 0.8564]$ with $\hat{f}^* = -1.0316$. Table 1 lists the detailed results at each iteration and Fig 3b-i show their progression after 3, 4, 5, and 10 iterations. SBOC identified the best solution correctly in the fifth iteration after 34 function evaluations.

**Table 1:** Iterative progression of SBOC (using RBF) for global minimization of Six Hump Camel Back Function

| $\boldsymbol{iter}$ | $\boldsymbol{K}$ | **Point Addition Strategy** | $\boldsymbol{x}$ | | $\boldsymbol{f}$ | $\hat{\boldsymbol{x}}^*$ | | $\hat{\boldsymbol{f}}^*$ |
|---|---|---|---|---|---|---|---|---|
| 0 | 10 | Initial Sampling by Sobol Technique | 0.5578 | 0.9748 | 0.0730 | 0.3853 | 0.8083 | -0.4732 |
| | | | 0.3233 | 0.1973 | 1.0156 | | | |
| | | | 0.8141 | 0.4830 | 2.3451 | | | |
| | | | 0.0483 | 0.6901 | 1.0924 | | | |

| $iter$ | $K$ | Point Addition Strategy | $\boldsymbol{x}$ | | $\boldsymbol{f}$ | $\widehat{\boldsymbol{x}}^*$ | | $\widehat{\boldsymbol{f}}^*$ |
|---|---|---|---|---|---|---|---|---|
| | | | 0.7448 | 0.0230 | 0.9367 | | | |
| | | | 0.3853 | 0.8083 | -0.4732 | | | |
| | | | 0.8752 | 0.5305 | 2.2416 | | | |
| | | | 0.2344 | 0.2999 | 2.2059 | | | |
| | | | 0.6171 | 0.3739 | 0.4236 | | | |
| | | | 0.2576 | 0.5810 | 1.9222 | | | |
| | 11 | Minimize $S(\boldsymbol{x})$ | 0 | 1 | 1.7333 | | | |
| 1 | 12 | Explore in between $C_1^* = 4$ clusters | 0.6810 | 0.1985 | 0.2050 | 0.4021 | 0.8590 | -0.7150 |
| | 13 | Exploit around $\widehat{\boldsymbol{x}}^*$ with $LN(\widehat{\boldsymbol{x}}^*) = 3$ | 0.4021 | 0.8590 | -0.7150 | | | |
| | 14 | Minimize $S(\boldsymbol{x})$ | 0.3964 | 0.8925 | -0.6445 | | | |
| 2 | 15 | Explore in between $C_1^* = 2$ clusters | 0.2460 | 0.4405 | 2.3231 | 0.4349 | 0.8854 | -0.9036 |
| | 16 | Exploit around $\widehat{\boldsymbol{x}}^*$ with $LN(\widehat{\boldsymbol{x}}^*) = 3$ | 0.4349 | 0.8854 | -0.9036 | | | |
| | 17 | Minimize $S(\boldsymbol{x})$ | 0.4498 | 0.8754 | -0.9765 | | | |
| 3 | 18 | Explore in between $C_1^* = 3$ clusters | 0.1530 | 0.6356 | 1.6463 | 0.4498 | 0.8754 | -0.9765 |
| | 19 | Exploit around $\widehat{\boldsymbol{x}}^*$ with $LN(\widehat{\boldsymbol{x}}^*) = 4$ | 0.4058 | 0.8627 | -0.7444 | | | |
| | 20 | Minimize $S(\boldsymbol{x})$ | 0.4574 | 0.8675 | -1.0045 | | | |
| 4 | 21 | Explore in between $C_1^* = 4$ clusters | 0.7156 | 0.4284 | 1.7470 | 0.4574 | 0.8675 | -1.0045 |
| | 22 | Exploit around $\widehat{\boldsymbol{x}}^*$ with $LN(\widehat{\boldsymbol{x}}^*) = 5$ | 0.4185 | 0.8751 | -0.8268 | | | |
| | 23 | Minimize $S(\boldsymbol{x})$ | 0.4755 | 0.8565 | -1.0314 | | | |
| 5 | 24 | Explore in between $C_1^* = 4$ clusters | 0.6490 | 0.2862 | 0.3185 | 0.4755 | 0.8565 | -1.0314 |
| | 25 | Exploit around $\widehat{\boldsymbol{x}}^*$ with $LN(\widehat{\boldsymbol{x}}^*) = 5$ | 0.4335 | 0.8732 | -0.9130 | | | |
| | 26 | Minimize $S(\boldsymbol{x})$ | 0.4742 | 0.8584 | -1.0308 | | | |
| 6 | 27 | Explore in between $C_1^* = 3$ clusters | 0.4861 | 0.2418 | -0.7415 | 0.4755 | 0.8565 | -1.0314 |
| | 28 | Exploit around $\widehat{\boldsymbol{x}}^*$ with $LN(\widehat{\boldsymbol{x}}^*) = 6$ | 0.4500 | 0.8705 | -0.9818 | | | |
| | 29 | Minimize $S(\boldsymbol{x})$ | 1.0000 | 0.0557 | 1.2915 | | | |
| 7 | 30 | Explore in between $C_1^* = 3$ clusters | 0.7129 | 0.1107 | 0.3055 | 0.4755 | 0.8565 | -1.0314 |
| | 31 | Exploit around $\widehat{\boldsymbol{x}}^*$ with $LN(\widehat{\boldsymbol{x}}^*) = 6$ | 0.4510 | 0.8712 | -0.9845 | | | |
| | 32 | Minimize $S(\boldsymbol{x})$ | 0.4776 | 0.8564 | -1.0316 | | | |
| 8 | 33 | Explore in between $C_1^* = 3$ clusters | 0.3214 | 0.6947 | 0.7460 | 0.4776 | 0.8564 | -1.0316 |
| | 34 | Exploit around $\widehat{\boldsymbol{x}}^*$ with $LN(\widehat{\boldsymbol{x}}^*) = 6$ | 0.4565 | 0.8672 | -1.0024 | | | |
| 9 | 35 | Explore in between $C_1^* = 3$ clusters | 0.2788 | 0.2486 | 1.6934 | 0.4776 | 0.8564 | -1.0316 |
| | 36 | Exploit around $\widehat{\boldsymbol{x}}^*$ with $LN(\widehat{\boldsymbol{x}}^*) = 7$ | 0.4594 | 0.8665 | -1.0093 | | | |
| 10 | 37 | Explore in between $C_1^* = 3$ clusters | 0.2402 | 0.3702 | 2.3088 | 0.4776 | 0.8564 | -1.0316 |
| 11 | 38 | Explore in between $C_1^* = 3$ clusters | 0.4047 | 0.2196 | -0.1108 | 0.4776 | 0.8564 | -1.0316 |
| | 39 | Exploit around $\widehat{\boldsymbol{x}}^*$ with $LN(\widehat{\boldsymbol{x}}^*) = 8$ | 0.4608 | 0.8657 | -1.0128 | | | |
| | 40 | Minimize $S(\boldsymbol{x})$ | 0.0000 | 0.0000 | 5.7333 | | | |
| 12 | 41 | Explore in between $C_1^* = 7$ clusters | 0.3533 | 0.7515 | 0.0904 | 0.4776 | 0.8564 | -1.0316 |
| | 42 | Exploit around $\widehat{\boldsymbol{x}}^*$ with $LN(\widehat{\boldsymbol{x}}^*) = 9$ | 0.4599 | 0.8663 | -1.0105 | | | |
| 13 | 43 | Explore in between $C_1^* = 3$ clusters | 0.5676 | 0.2640 | -0.5391 | 0.4776 | 0.8564 | -1.0316 |
| | 44 | Exploit around $\widehat{\boldsymbol{x}}^*$ with $LN(\widehat{\boldsymbol{x}}^*) = 9$ | 0.4607 | 0.8654 | -1.0128 | | | |
| 14 | 45 | Explore in between $C_1^* = 3$ clusters | 0.2518 | 0.5107 | 2.1986 | 0.4776 | 0.8564 | -1.0316 |
| | 46 | Exploit around $\widehat{\boldsymbol{x}}^*$ with $LN(\widehat{\boldsymbol{x}}^*) = 9$ | 0.4618 | 0.8649 | -1.0150 | | | |
| 15 | 47 | Explore in between $C_1^* = 3$ clusters | 0.1394 | 0.1243 | 2.3345 | 0.4776 | 0.8564 | -1.0316 |
| 16 | 48 | Explore in between $C_1^* = 4$ clusters | 0.4454 | 0.2307 | -0.5204 | 0.4776 | 0.8564 | -1.0316 |
| | 49 | Exploit around $\widehat{\boldsymbol{x}}^*$ with $LN(\widehat{\boldsymbol{x}}^*) = 10$ | 0.4626 | 0.8644 | -1.0167 | | | |
| 17 | 50 | Explore in between $C_1^* = 3$ clusters | 0.2895 | 0.6378 | 1.3861 | 0.4776 | 0.8564 | -1.0316 |
| | 51 | Exploit around $\widehat{\boldsymbol{x}}^*$ with $LN(\widehat{\boldsymbol{x}}^*) = 10$ | 0.4631 | 0.8642 | -1.0176 | | | |

## 5. Performance Assessment Strategy

We now assess the performance of SBOC on 52 diverse analytical test functions by comparing against several existing optimization algorithms.

### 5.1 Analytical Test Functions

Table 2 lists the 52 continuous and differentiable test functions (TFs) with varying complex shapes and $N$ ranging from 2 to 10. Table S1 presents their analytical expressions, shape profiles, reference sources (Jamil and Yang, 2013; Surjanovic and Bingham, 2013), and summary features. These TFs have been widely used in the literature for assessing optimization algorithms. 46 TFs are multimodal and non-convex, while the rest are unimodal and convex. For 16 TFs marked by * in Table 2, a global optimum is exactly at the center of the domain (i.e. $\boldsymbol{x}^* = [0.5]^N$). We call them midpoint-optimal test functions (MOTFs). $\boldsymbol{x}^*$ and $f^*$ are also listed for all TFs in Table 2.

**Table 2:** List of test functions along with their $N$, bounds, $\boldsymbol{x}^*$ and $f^*$

| TF | Function Name | $N$ | Bounds | $\boldsymbol{x}^*$ | $\boldsymbol{f}^*$ |
|---|---|---|---|---|---|
| 1 | Six Hump Camel Back | 2 | $[-2,2]$, $[-1,1]$ | $[0.5225, 0.1437]$, $[0.4775, 0.8563]$ | $-1.0316$ |
| 2 | Ackley3 | 2 | $[-32,32]^2$ | $[0.4893, 0.4944]$ | $-195.629$ |
| 3 | Ackley4 | 2 | $[-5,5]^2$ | $[0.3490, 0.4245]$ | $-4.5901$ |
| 4 | Beale | 2 | $[-4.5,4.5]^2$ | $[0.8333, 0.5556]$ | 0 |
| 5 | Branin | 2 | $[-5,10]$, $[0,15]$ | $[0.1239, 0.8183]$, $[0.5428, 0.1517]$, $[0.9617, 0.1650]$ | 0.3979 |
| 6 | Cross in Tray | 2 | $[-10,10]^2$ | $[0.5697, 0.4303]$, $[0.5697, 0.5697]$, $[0.4303, 0.5697]$, $[0.4303, 0.4303]$ | $-2.0626$ |
| 7 | Easom | 2 | $[-10,10]^2$ | $[0.6571, 0.6571]$ | $-1$ |
| 8 | Eggholder | 2 | $[-512,512]^2$ | $[1, 0.8948]$ | $-959.641$ |
| 9 | Goldstein Price | 2 | $[-2,2]^2$ | $[0.5, 0.25]$ | 3 |
| 10 | Holder Table | 2 | $[-10,10]^2$ | $[0.9028, 0.9832]$, $[0.9028, 0.0168]$, $[0.0972, 0.9832]$, $[0.0972, 0.0168]$ | $-19.2085$ |
| 11 | Michalewicz | 2 | $[0,\pi]^2$ | $[0.7002, 0.4997]$ | $-1.8013$ |
| 12 | Schwefel | 2 | $[-500,500]^2$ | $[0.9210, 0.9210]$ | 0 |
| 13 | Shubert | 2 | $[-5.12,5.12]^2$ | $[0.3608, 0.4218]$, $[0.4218, 0.3608]$, $[0.4218, 0.9744]$, $[0.9744, 0.4218]$ | $-186.731$ |
| 14 | Styblinski Tang | 2 | $[-5,5]^2$ | $[0.2096, 0.2096]$ | $-78.332$ |
| 15 | McCormick | 2 | $[-1.5,4]$, $[-3,4]$ | $[0.1732, 0.2075]$ | $-1.9133$ |
| 16 | Hartmann3 | 3 | $[0,1]^3$ | $[0.1146, 0.5556, 0.8525]$ | $-3.8628$ |
| 17 | Shekel5 | 4 | $[0,10]^4$ | $[0.4, 0.4, 0.4, 0.4]$ | $-10.1532$ |
| 18 | Shekel7 | 4 | $[0,10]^4$ | $[0.4, 0.4, 0.4, 0.4]$ | $-10.4029$ |
| 19 | Trid | 5 | $[-25,25]^5$ | $[0.6, 0.66, 0.68, 0.66, 0.6]$ | $-30$ |
| 20 | Hartmann6 | 6 | $[0,1]^6$ | $[0.2017, 0.1500, 0.4769, 0.2753, 0.3117, 0.6573]$ | $-3.0425$ |
| 21 | Bukin | 2 | $[-15,-5]$, $[-3,3]$ | $[0.5, 0.6667]$ | 0 |
| 22* | Griewank | 5 | $[-600,600]^5$ | $[0.5, 0.5, 0.5, 0.5, 0.5]$ | 0 |
| 23 | Levy | 6 | $[-10,10]^6$ | $[0.55, 0.55, 0.55, 0.55, 0.55, 0.55]$ | 0 |
| 24 | Levy13 | 2 | $[-10,10]^2$ | $[0.55, 0.55]$ | 0 |
| 25* | Rastrigin | 6 | $[-5.12,5.12]^6$ | $[0.5, 0.5, 0.5, 0.5, 0.5, 0.5]$ | 0 |
| 26 | Perm | 5 | $[-5,5]^5$ | $[0.6, 0.7, 0.8, 0.9, 1]$ | 0 |

| TF | Function Name | $N$ | Bounds | $\boldsymbol{x}^*$ | $\boldsymbol{f}^*$ |
|---|---|---|---|---|---|
| 27* | Sum of Squares | 4 | $[-5.12, 5.12]^4$ | $[0.5, 0.5, 0.5, 0.5]$ | 0 |
| 28 | Booth | 2 | $[-10, 10]^2$ | $[0.55, 0.65]$ | 0 |
| 29 | Rosenbrock | 3 | $[-2.048, 2.048]^3$ | $[0.7441, 0.7441, 0.7441]$ | 0 |
| 30* | Griewank | 2 | $[-50, 50]^2$ | $[0.5, 0.5]$ | 0 |
| 31* | Rastrigin | 2 | $[-5.12, 5.12]^2$ | $[0.5, 0.5]$ | 0 |
| 32 | Perm | 2 | $[-2, 2]^2$ | $[0.75, 1]$ | 0 |
| 33 | Perm | 3 | $[-3, 3]^3$ | $[0.6667, 0.8333, 1]$ | 0 |
| 34 | Adjiman | 2 | $[-1, 2], [-1, 1]$ | $[1, 0.5529]$ | $-2.0218$ |
| 35* | Alpine | 2 | $[-10, 10]^2$ | $[0.5, 0.5]$ | 0 |
| 36* | Alpine | 4 | $[-10, 10]^4$ | $[0.5, 0.5, 0.5, 0.5]$ | 0 |
| 37* | Alpine | 6 | $[-10, 10]^6$ | $[0.5, 0.5, 0.5, 0.5, 0.5, 0.5]$ | 0 |
| 38* | Bartels Conn | 2 | $[-500, 500]^2$ | $[0.5\ 0.5]$ | 1 |
| 39 | Bird | 2 | $[-6.284, 6.284]^2$ | $[0.8740, 0.7509], [0.3741, 0.2508]$ | $-106.765$ |
| 40 | Colville | 4 | $[-10, 10]^4$ | $[0.55, 0.55, 0.55, 0.55]$ | 0 |
| 41 | Dixon and Price | 2 | $[-10, 10]^2$ | $[0.55, 0.5354]$ | 0 |
| 42 | Dixon and Price | 4 | $[-10, 10]^4$ | $[0.55, 0.5353, 0.5297, 0.5273]$ | 0 |
| 43* | Exponential | 2 | $[-1, 1]^2$ | $[0.5, 0.5]$ | $-1$ |
| 44 | Hosaki | 2 | $[0, 5], [0, 6]$ | $[0.8, 0.3333]$ | $-2.3458$ |
| 45 | Miele Cantrell | 4 | $[-1, 1]^4$ | $[0.5, 1, 1, 1]$ | 0 |
| 46* | Price | 2 | $[-10, 10]^2$ | $[0.5, 0.5]$ | 0.9 |
| 47* | Salomon | 3 | $[-100, 100]^3$ | $[0.5, 0.5, 0.5]$ | 0 |
| 48* | Ackley | 6 | $[-5, 5]^6$ | $[0.5, 0.5, 0.5, 0.5, 0.5, 0.5]$ | 0 |
| 49* | Exponential | 6 | $[-1, 1]^6$ | $[0.5, 0.5, 0.5, 0.5, 0.5, 0.5]$ | $-1$ |
| 50 | Schwefel | 10 | $[0,10]^{10}$ | $[0.1, 0.1, 0.1, 0.1, 0.1, 0.1, 0.1, 0.1, 0.1, 0.1]$ | 0 |
| 51* | Wavy | 10 | $[-\pi, \pi]^{10}$ | $[0.5, 0.5, 0.5, 0.5, 0.5, 0.5, 0.5, 0.5, 0.5, 0.5]$ | 0 |
| 52* | Zakharov | 10 | $[-5,5]^{10}$ | $[0.5, 0.5, 0.5, 0.5, 0.5, 0.5, 0.5, 0.5, 0.5, 0.5]$ | 0 |

**NOTE:** $\boldsymbol{x}^*$ is defined for the normalized domain $[0, 1]^N$.

*MOTFs with $\boldsymbol{x}^*$ at the center of domain.

## 5.2 $\tilde{S}(\boldsymbol{x})$ in SBOC

Although SBOC is agnostic to the choice of the surrogate modeling technique, it is advisable to employ a flexible $\tilde{S}(\boldsymbol{x})$ capable of accurately capturing the objective function profile to effectively navigate the search towards a global optimum. Therefore, we consider two different surrogate forms of high-fidelity, an RBF form with multi-quadratic basis function $\tau_k(\boldsymbol{x})$ and linear tail function $t(\boldsymbol{x})$ (Eq. 8) and a KRG form with Gaussian correlation function and a second-degree polynomial regression function, to implement SBOC.

KRG modeling technique uses a regression function to learn the underlying trend in response and a correlation function to capture the error or lack of fit at each train point, as shown below,

$$\tilde{S}(\boldsymbol{x}) = reg(\boldsymbol{x}) + \sum_{k=1}^{K} \gamma_k c\left[\boldsymbol{x}, \boldsymbol{x}^{(k)}, \boldsymbol{\theta}\right] \tag{9}$$

where, $reg(\boldsymbol{x})$ denotes the regression function, $c$ denotes the correlation function, and $\boldsymbol{\theta}$ denotes the $N$ hyperparameters determined by maximum likelihood estimation technique, and the parameters in $reg(\boldsymbol{x})$ and $\gamma_k, k = 1, 2, \dots, K$ are the model parameters.

We use SBOC-R to refer to the SBOC implementation with the RBF surrogate form and SBOC-K to denote its implementation using the KRG form.

## 5.3 Benchmarking Algorithms

We compare SBOC-R and SBOC-K with sixteen benchmarking algorithms. These are Generalized Pattern Search (GPS) of (Torczon, 1997), Mesh Adaptive Direct Search (MADS) of (Audet and Dennis, 2006), Implicit Filtering (IMPFIL) of (Gilmore and Kelley, 1995), GLOBAL of (Csendes et al., 2008), DIRECT-G of (Stripinis et al., 2018), Multilevel Coordinate Search (MCS) of (Huyer and Neumaier, 1999), Genetic Algorithm (GA) of (Fraser, 1958), Simulated Annealing (SA) of (Kirkpatrick, 1984), Particle Swarm Optimization (PSO) of (Kennedy and Eberhart, 1995), Bayesian Adaptive Direct Search (BADS) of (Acerbi and Ji, 2017), Stable Noisy Optimization by Branch and Fit (SNOBFIT) of (Huyer and Neumaier, 2008), Costly Global Optimization (CGO) of (Gutmann, 2001), Multistart Local Metric Stochastic Response Surface (MLMSRS) of (Regis and Shoemaker, 2007a), Two Stage Framework (TSF) of (Garud et al., 2019), Efficient Global Optimization (EGO) of (Jones et al., 1998), and Branch-and-Model (BAM) of (Ma et al., 2023). These algorithms collectively represent a comprehensive set of state-of-the art DFO algorithms with differing methodologies. GPS, MADS, IMPFIL, GLOBAL, DIRECT-G, and MCS are direct search algorithms; GA, SA, and PSO are metaheuristics approaches; and BADS, SNOBFIT, CGO, MLMSRS, TSF, EGO, and BAM are SBO algorithms. While some algorithms such as GPS, MADS, GA, SA, PSO, CGO, EGO are widely used for global optimization due to their simplicity, familiarity, and robustness, others have showcased strong performance in their respective studies. Many of these algorithms have also been used as benchmarks in recent works (Acerbi and Ji, 2017; Garud et al., 2019; Ma et al., 2023; Srinivas and Karimi, 2024) proposing novel DFO algorithms.

All algorithms except EGO and BAM were implemented in MATLAB R2023b. EGO and BAM were implemented in Python 3.8. The system was equipped with 32-core 2.90 GHz Intel CPU with 128 GB RAM. While full details of each algorithm are in their source reference, some aspects that may influence our assessment and inferences are as follows.

1. IMPFIL does not support termination based on $K_{max}$. It terminates when no better points exist in the neighborhood of $\hat{\boldsymbol{x}}_i^*$ (stencil failure) or $\hat{f}_i^*$ improves only marginally for a few consecutive iterations.
2. GLOBAL also does not support termination based on $K_{max}$. It terminates when no new basins of attractions exist. It also does not save its iteration history (i.e. $\hat{\boldsymbol{x}}_i^*$ and $\hat{f}_i^*$).
3. CGO is used from the free-licensed version of TOMLAB's optimization toolbox, which limits applicability to TFs with $N \leq 4$.
4. TSF could not be used with $N > 6$ due to its excessive computational demand for constructing Delaunay triangulations.

5. DIRECT-G is a deterministic algorithm that always begins its search from the center of $D$. Thus, it is certain to identify the global minimum for each MOTF.
6. MLMSRS resets its search with new initial points, when it stagnates for some consecutive iterations. During the first reset, it adds a new point at the center of $D$. This guarantees its success in locating the global minimum for each MOTF.
7. BAM always evaluates the center of $D$ in its first iteration as observed from the iterative history across all runs. Therefore, like DIRECT-G, it is guaranteed to successfully identify the global minimum for each MOTF.

### 5.4 Performance Metrics

We measure algorithmic performance by (1) $\Delta x^*$ or proximity of $\hat{\boldsymbol{x}}^*$ to its nearest global minimum $\boldsymbol{x}^*$, (2) $\Delta f^*$ or deviation of $\hat{f}^*$ from its nearest global minimum $f^*$, and (3) $\gamma$ or computational effort measured in terms of $f(\boldsymbol{x})$ evaluations.

$$\Delta x^* = \min_{\boldsymbol{x}^*} \|\hat{\boldsymbol{x}}^* - \boldsymbol{x}^*\| / \sqrt{N} \qquad 0 \le \Delta x^* \le 1 \tag{10}$$

$$\Delta f^* = \begin{cases} (\hat{f}^* - f^*)/|f^*|, & if\ f^* \neq 0 \\ \min(1, \hat{f}^*), & if\ f^* = 0 \end{cases} \qquad 0 \le \Delta f^* \le 1 \tag{11}$$

$$\gamma = \min[K^*, K_{max}] / K_{max} \qquad 0 \le \gamma \le 1 \tag{12}$$

$K^*$ in Eq. 12 denotes the number of function evaluations needed by an algorithm to achieve $\Delta f^* \le \Delta f^{**}$, where $\Delta f^{**}$ is some user-specified threshold. For GLOBAL, $K^*$ is its number of $f(\boldsymbol{x})$ evaluations at termination, because we do not know its history. If IMPFIL and GLOBAL terminate before $K_{max}$ without reaching $\Delta f^* \le \Delta f^{**}$, then we take $K^* = K_{max}$.

Lower values are better for all three measures.

### 5.5 Assessment Protocol

For all algorithms except IMPFIL and GLOBAL, we set $K_{max} = 100N$ and $\Delta f^{**} = 0.01$. For SBOC-R, we use $\tilde{S}(\boldsymbol{x})$ defined by Eq. 8, while for SBOC-K, we use $\tilde{S}(\boldsymbol{x})$ defined by Eq. 9. Apart from DIRECT-G, all algorithms are affected by the choice of initial point(s). Thus, we run each algorithm (except DIRECT-G and EGO) ten times for each TF with random initializations. DIRECT-G, being deterministic, is run only once. EGO is run five times, each starting from a different set of randomly sampled $5N$ initial points, owing to its long computational time requirements for executing high-dimensional test functions. From the performances over all runs for each TF, we find the medians ($\widetilde{\Delta x}$, $\widetilde{\Delta f}^*$, and $\tilde{\gamma}$) and use them as their indicative performances. We define success when an algorithm's indicative performance meets the desired solution quality, i.e. $\widetilde{\Delta f}^* \le \Delta f^{**}$. Since DIRECT-G, MLMSRS, and BAM will surely succeed for the 16 MOTFs, we compare SBOC with these three algorithms on the 36 non-MOTFs only to be fair and unbiased. All other algorithms are compared with SBOC-R and SBOC-K on all 52 TFs.

We consider the following aspects in comparing SBOC-R and SBOC-K with other algorithms.

1. Success rate ($S$) of an algorithm = Percentage of TFs on which it succeeds.
2. Averages of indicative performances ($\widetilde{\Delta x}$, $\widetilde{\Delta f}^*$, and $\tilde{\gamma}$) of an algorithm over all TFs.
3. Variations of an algorithm's indicative performances with $N$.
4. Variations of an algorithm's indicative performances with TF complexity and shape profile.

In addition to our analyses based on indicative performance, we also compare the average wall-clock times required by each algorithm to complete one run for all test functions.

## 6. Results and Discussion

We now analyze the performance of SBOC-R and SBOC-K using the assessment protocol discussed above.

### 6.1 Success Rate

Table 3 lists $S$ for various algorithms. Our key findings are as follows:

1. SBOC-K and MCS achieved the highest $S = 65.4\%$, while SBOC-R and BADS achieved the second highest $S = 63.5\%$. All others (excluding DIRECT-G, MLMSRS, and BAM) had $S < 58\%$ on all 52 TFs. For non-MOTFs, SBOC-K with $S = 80.6\%$ and SBOC-R with $S = 75\%$ outperformed DIRECT-G ($S = 63.9\%$), MLMSRS ($S = 63.9\%$), and BAM ($S = 58.3\%$).
2. 4/5 of the top ranked algorithms are SBO algorithms, namely SBOC-K, SBOC-R, BADS, and SNOBFIT. This demonstrates the ability of SBO algorithms to effectively capture the unknown $f(\boldsymbol{x})$ landscape and guide the search towards a global optimum.
3. Despite being a direct search algorithm, MCS succeeded on the maximum number of test functions. An efficient exploration-exploitation strategy based on systematic splitting of the boxed domain into smaller boxes combined with recursive local searches guided by local quadratic models enables it to identify a global optimum for complex test functions.
4. 4/5 of the top ranked algorithms (SBOC-K, MCS, SBOC-R, and SNOBFIT) claim to be global search methods, hence their success seems natural. However, BADS, despite being a local SBO algorithm, performed the same as SBOC-R. This can be accredited to its effective, dynamic exploration strategy based on Bayesian optimization.
5. Among the direct search algorithms, GLOBAL had the lowest $S$ of 42.3%, despite employing a global search strategy.
6. Metaheuristics, despite their frequent claims of being global optimizers, performed the poorest of all. GA achieved $S = 17.2\%$, SA achieved $S = 30.8\%$, and PSO achieved $S = 28.8\%$.

**Table 3:** Success Rate for each algorithm

| Algorithm | $S$ over all TFs |
|---|---|
| SBOC-R | 63.5% |
| SBOC-K | 65.4% |
| GPS | 53.8% |
| MADS | 48.1% |
| IMPFIL | 51.9% |
| GLOBAL | 42.3% |
| MCS | 65.4% |
| GA | 17.3% |
| SA | 30.8% |
| PSO | 28.8% |
| BADS | 63.5% |
| SNOBFIT | 57.7% |
| CGO | 51.9% |
| TSF | 46.1% |
| EGO | 44.2% |
| **Algorithm** | **$S$ over non-MOTFs** |
| SBOC-R | 75% |
| SBOC-K | 80.6% |
| DIRECT-G | 63.9% |
| MLMSRS | 63.9% |
| BAM | 58.3% |

## 6.2 Overall Solution Quality

Table S2 and S3 in the supplementary material reports the indicative solution qualities ($\widetilde{\Delta x}$ and $\widetilde{\Delta f^*}$) for each TF and algorithm. We define overall solution qualities as averages ($\overline{\Delta x}$ and $\overline{\Delta f^*}$) of indicative qualities ($\widetilde{\Delta x}$ and $\widetilde{\Delta f^*}$) over all TFs for each algorithm. Table 4 lists $\overline{\Delta x}$ and $\overline{\Delta f^*}$ for all algorithms.

SNOBFIT (0.031), SBOC-K (0.042), MADS (0.045), GPS (0.046), PSO (0.049), and SBOC-R (0.050) achieved the lowest $\overline{\Delta x}$. The ability of SNOBFIT to approach $\boldsymbol{x}^*$ more accurately can be attributed to two key factors. First, it constructs multiple local quadratic surrogates and minimizes each surrogate to add all distinct minimizers at each iteration. This strategy allows it to capture the local curvature of $f(\boldsymbol{x})$ more effectively than single global surrogates. Second, its termination tolerances (step size, optimality, and $f(\boldsymbol{x})$ tolerance) in MINQ (Matlab program for bound constrained Indefinite Quadratic programming) (Huyer and Neumaier, 2018) are very stringent. This allowed SNOBFIT to converge exactly to $\boldsymbol{x}^*$ for TF3, TF19, TF27, TF28, TF32, TF38, TF43, and TF49 (Table S2). The two implementations of SBOC, SBOC-K and SBOC-R were among the top six algorithms based on $\overline{\Delta x}$. The $\overline{\Delta x}$ of SBOC-K, SBOC-R, MADS, GPS, and PSO were quite comparable. For non-MOTFs, SBOC-K with $\overline{\Delta x} = 0.028$ and SBOC-R with $\overline{\Delta x} = 0.046$ outperformed MLMSRS (0.087), DIRECT-G (0.092), and BAM (0.092) considerably.

CGO (0.133), MCS (0.159), TSF (0.184), SBOC-R (0.191), SNOBFIT (0.192), and SBOC-K (0.200) achieved the lowest $\overline{\Delta f^*}$. The success of CGO and TSF is largely due to the small TFs ($N \leq 4$ for CGO and $N \leq 6$ for TSF), on which they are tested. While MCS did not rank among the top six algorithms based on $\overline{\Delta x}$, it achieved the second best $\overline{\Delta f^*}$ underlining its ability to determine the global minimum, albeit without consistently converging to $\boldsymbol{x}^*$ precisely for several test functions. Moreover, large $\widetilde{\Delta x} > 0.15$ for 9/18 TFs (TF8, TF12, Tf17, TF18, TF21, TF26, TF33, TF51, and TF52, see Table S2 in supplementary file) for which MCS failed (i.e. $\widetilde{\Delta f^*} \leq \Delta f^{**} = 0.01$) led to a higher $\overline{\Delta x}$ over all TFs. SNOBFIT also achieved a low $\overline{\Delta f^*}$ due to its features discussed before. Like $\overline{\Delta x}$, SBOC-K and SBOC-R were among the top six algorithms based on $\overline{\Delta f^*}$. For non-MOTFs also, SBOC-K (0.132) and SBOC-R (0.151) outperformed MLMSRS (0.173), BAM (0.178), and DIRECT-G (0.183).

Other findings on overall solution quality are listed below.

1. 5/6 of the top ranked algorithms based on $\overline{\Delta f^*}$ are SBO algorithms.

2. CGO, MCS, and TSF do well on $\overline{\Delta f^*}$ but not on $\overline{\Delta x}$.

3. EGO faired poorly based on both $\overline{\Delta x}$ and $\overline{\Delta f^*}$.

4. Metaheuristics were the poorest among all algorithms on $\overline{\Delta f^*}$.

5. The large $S$ for BADS did not translate into low $\overline{\Delta x}$ or $\overline{\Delta f^*}$. BADS did not feature among the top five algorithms on either metric. Seemingly, BADS is able to find global optima in many instances, but it gets stuck in some local optima due to its local search, which then impacts these two metrics drastically.

**Table 4:** Overall solution quality based on $\overline{\Delta x}$ and $\overline{\Delta f^*}$ for each algorithm

| Algorithm | $\overline{\Delta x}$ over all TFs | $\overline{\Delta f^*}$ over all TFs |
|---|---|---|
| SBOC-R | 0.050 | 0.191 |
| SBOC-K | 0.042 | 0.200 |
| GPS | 0.046 | 0.201 |
| MADS | 0.045 | 0.225 |
| IMPFIL | 0.081 | 0.268 |
| GLOBAL | 0.080 | 0.476 |
| MCS | 0.066 | 0.159 |
| GA | 0.088 | 0.450 |
| SA | 0.096 | 0.392 |
| PSO | 0.049 | 1.857 |
| BADS | 0.061 | 0.339 |
| SNOBFIT | 0.031 | 0.192 |

| Algorithm | $\overline{\Delta x}$ over all TFs | $\overline{\Delta f^*}$ over all TFs |
|---|---|---|
| CGO | 0.054 | 0.133 |
| TSF | 0.078 | 0.184 |
| EGO | 0.058 | 0.272 |
| **Algorithm** | $\overline{\Delta x}$ **over non-MOTFs** | $\overline{\Delta f^*}$ **over non-MOTFs** |
| SBOC-R | 0.046 | 0.151 |
| SBOC-K | 0.028 | 0.132 |
| DIRECT-G | 0.092 | 0.183 |
| MLMSRS | 0.087 | 0.173 |
| BAM | 0.092 | 0.178 |

### 6.3 Overall Computational Effort

As done for solution quality, we define overall compute effort ($\bar{\gamma}$) as the average of indicative compute efforts ($\tilde{\gamma}$) on all TFs for each algorithm. Table S4 of the supplementary material lists $\tilde{\gamma}$ for each TF and algorithm. Table 5 lists $\bar{\gamma}$ for all algorithms.

BADS (0.49), SBOC-K (0.50), CGO (0.53), SBOC-R (0.54), and TSF (0.57) demanded the least compute efforts. While CGO and TSF enjoyed the advantage of smaller TFs, both SBOC implementations (SBOC-R and SBOC-K) performed well in comparison. Low $\bar{\gamma}$ for BADS shows the advantage of using Bayesian optimization within a direct search framework. Even on the non-MOTFs, SBOC-K (0.40) and SBOC-R (0.48) significantly outperformed MLMSRS (0.52), BAM (0.57), and DIRECT-G (0.64).

SBO algorithms in general outperformed others based on $\bar{\gamma}$, with the top five algorithms being SBO methods. However, EGO required larger compute effort than some direct search algorithms such as IMPFIL and MCS. Metaheuristics demanded the maximum compute efforts.

**Table 5:** Computational Effort for each algorithm

| Algorithm | $\bar{\gamma}$ over all TFs |
|---|---|
| SBOC-R | 0.54 |
| SBOC-K | 0.50 |
| GPS | 0.84 |
| MADS | 0.84 |
| IMPFIL | 0.61 |
| GLOBAL | 0.74 |
| MCS | 0.64 |
| GA | 0.93 |
| SA | 0.89 |
| PSO | 0.90 |
| BADS | 0.49 |
| SNOBFIT | 0.64 |
| CGO | 0.53 |
| TSF | 0.57 |
| EGO | 0.68 |
| **Algorithm** | $\bar{\gamma}$ **over non-MOTFs** |
| SBOC-R | 0.48 |
| SBOC-K | 0.40 |
| DIRECT-G | 0.64 |
| MLMSRS | 0.52 |
| BAM | 0.57 |

### 6.4 Performance Variations with *N*

TF size (or $N$) has much impact on algorithmic performance. The search space $D$ increases exponentially with $N$. Boukouvala and Floudas, 2017 and Rios and Sahinidis, 2013 have numerically demonstrated the *curse of dimensionality* on the quality and reliability of solutions from various algorithms. Our study (Table 2) comprised 40 low-dimensional ($N \leq 4$) and 12 high-dimensional ($N > 4$) TFs. Table 6 lists $S$, $\overline{\Delta x}$, $\overline{\Delta f^*}$, and $\bar{\gamma}$ for all algorithms on these two classes of TFs.

On TFs with $N \leq 4$, SBOC-K achieved the best $S = 70\%$, while SBOC-R was the third best with $S = 65\%$ along with SNOBFIT, BADS, and GPS after CGO and MCS with $S = 67.5\%$. SBOC-K along with MADS ranked second based on $\overline{\Delta x}$ with 0.041 trailing SNOBFIT (0.026). SBOC-R ranked fifth with 0.051, whereas GPS (0.042) and PSO (0.057) ranked third and fourth, respectively. Based on $\overline{\Delta f^*}$, SBOC-K (0.157) and SBOC-R (0.160) ranked seventh and eighth respectively, behind GPS (0.117), SNOBFIT (0.121), MCS (0.127), CGO (0.133), TSF (0.145), and MADS (0.148). SBOC-K achieved the best $\bar{\gamma} = 0.460$, followed by BADS (0.467), CGO (0.532), and SBOC-R (0.534). Both SBOC-K and SBOC-R outperformed DIRECT-G, MLMSRS, and BAM based on $S$, $\overline{\Delta x}$, and $\bar{\gamma}$ on the 31 non-MOTFs with $N \leq 4$. However, based on $\overline{\Delta f^*}$, BAM (0.1425) was marginally than SBOC-R (0.143), whereas SBOC-K (0.1216) outperformed BAM.

On TFs with $N > 4$, SBOC-R, MCS, and BADS achieved the highest $S = 58.3\%$, followed by SBOC-K and IMPFIL, each with $S = 50\%$. SBOC-K and SBOC-R were the best based on $\overline{\Delta x}$ with 0.043 and 0.046, respectively. BADS achieved the lowest $\overline{\Delta f^*} = 0.187$, followed by MCS (0.264), SBOC-R (0.295) and SBOC-K (0.345). In terms of $\bar{\gamma}$, while SBOC-R (0.563) required the least compute effort, SBOC-K (0.621) ranked fourth after BADS (0.585) and TSF (0.620). Both SBOC-K and SBOC-R outperformed DIRECT-G, MLMSRS, and BAM in all four measures on the five non-MOTFs with $N > 4$. Clearly, SBOC is better for larger, more difficult problems.

**Table 6:** Algorithm performances over low-dimensional and high-dimensional TFs

| **Algorithm** | **Low-dimensional TFs** | | | | **High-dimensional TFs** | | | |
|---|---|---|---|---|---|---|---|---|
| | $S$ | $\overline{\Delta x}$ | $\overline{\Delta f^*}$ | $\bar{\gamma}$ | $S$ | $\overline{\Delta x}$ | $\overline{\Delta f^*}$ | $\bar{\gamma}$ |
| SBOC-R | 65% | 0.051 | 0.160 | 0.534 | 58.3% | 0.046 | 0.295 | 0.563 |
| SBOC-K | 70% | 0.041 | 0.157 | 0.460 | 50% | 0.043 | 0.345 | 0.621 |
| GPS | 65% | 0.042 | 0.117 | 0.800 | 16.7% | 0.059 | 0.480 | 0.962 |
| MADS | 60% | 0.041 | 0.148 | 0.799 | 8.3% | 0.057 | 0.480 | 0.987 |
| IMPFIL | 52.5% | 0.081 | 0.226 | 0.605 | 50% | 0.077 | 0.407 | 0.636 |
| GLOBAL | 45% | 0.080 | 0.492 | 0.667 | 33.3% | 0.078 | 0.426 | 1 |
| MCS | 67.5% | 0.066 | 0.127 | 0.633 | 58.3% | 0.063 | 0.264 | 0.656 |
| GA | 22.5% | 0.084 | 0.380 | 0.908 | 0% | 0.099 | 0.685 | 1 |
| SA | 40% | 0.075 | 0.288 | 0.852 | 0% | 0.168 | 0.740 | 1 |
| PSO | 35% | 0.047 | 2.227 | 0.874 | 8.3% | 0.058 | 0.625 | 0.996 |
| BADS | 65% | 0.065 | 0.384 | 0.467 | 58.3% | 0.047 | 0.187 | 0.585 |
| SNOBFIT | 65% | 0.026 | 0.121 | 0.593 | 33.3% | 0.048 | 0.426 | 0.781 |
| CGO | 67.5% | 0.054 | 0.133 | 0.532 | - | - | - | - |
| TSF | 52.5% | 0.083 | 0.145 | 0.561 | 25% | 0.059 | 0.359 | 0.620 |
| EGO | 55% | 0.033 | 0.158 | 0.611 | 8.3% | 0.145 | 0.654 | 0.928 |
| **Algorithm** | **Low-dimensional non-MOTFs** | | | | **High-dimensional non-MOTFs** | | | |
| | $S$ | $\overline{\Delta x}$ | $\overline{\Delta f^*}$ | $\bar{\gamma}$ | $S$ | $\overline{\Delta x}$ | $\overline{\Delta f^*}$ | $\bar{\gamma}$ |
| SBOC-R | 74.2% | 0.047 | 0.1430 | 0.479 | 80% | 0.042 | 0.200 | 0.471 |
| SBOC-K | 80.6% | 0.028 | 0.1216 | 0.395 | 80% | 0.029 | 0.200 | 0.455 |
| DIRECT-G | 71.0% | 0.082 | 0.1432 | 0.604 | 20% | 0.156 | 0.432 | 0.896 |
| MLMSRS | 67.7% | 0.077 | 0.1573 | 0.496 | 20% | 0.147 | 0.270 | 0.678 |
| BAM | 58.1% | 0.085 | 0.1425 | 0.577 | 60% | 0.129 | 0.400 | 0.501 |
| **NOTE:** CGO cannot be run on high-dimensional TFs, hence its metrics are not defined | | | | | | | | |

## 6.5 Performance Variations with Convexity and Shape Profile

We analyze algorithmic performances on unimodal vs multimodal TFs and on a few TFs with unique shape profiles.

TF19, TF27, TF28, TF29, TF41, TF42, TF52 are unimodal and convex. SBOC-K successfully identified the global minimum for all these except TF52, where SBOC-R was successful for all except TF42 and TF52. These test functions are scalable to any $N$, so we show their surface plots for $N = 2$ in Figs. 4a and 4b respectively. TF42 (Dixon and Price Function with $N = 4$) is a non-MOTF with a long, narrow, flat valley. Its global minimum lies on the valley's floor. The valley's small curvature makes convergence to $\boldsymbol{x}^*$ particularly difficult. Only SBOC-K, IMPFIL, CGO, TSF, and BAM optimized TF42 successfully. TF52 (Zakharov Function with $N = 10$) has a plate-shaped profile with the global minimum at the domain center on a nearly flat surface. Only DIRECT-G, MLMSRS, and BAM could reach this minimum due to their inherent bias towards the domain center. All other unimodal TFs except TF27 also have their global optima in nearly flat regions. However, both SBOC-K and SBOC-R successfully optimized all of them within $K_{max}$ evaluations.

Among the 45 multimodal and non-convex TFs, no algorithm succeeded on TF21 (Bukin with $N = 2$), TF26 (Perm with $N = 5$) and TF40 (Colville with $N = 4$). All had $\bar{\gamma} = 1$, indicating that they all needed more $f(\boldsymbol{x})$ evaluations to reach a global optimum. TF21 (Fig. 4c) has several local minima along a narrow and long ridge, which makes convergence to the global minimum challenging. TF26 (Fig. 4d for its 2D surface plot) has local minima embedded within a flat interior basin. TF40 has several local optima, but it is non-scalable, so cannot be plotted or visualized.

Excluding DIRECT-G, MLMSRS, and BAM, no algorithm succeeded on some complex multimodal MOTFs such as TF22 (Griewank with $N = 5$), TF25 (Rastrigin with $N = 6$), and TF47 (Salomon with $N = 3$). All had $\bar{\gamma} = 1$ for these TFs, again indicating the need for more computational effort. Additionally, TF30 (Griewank with $N = 2$) could only be optimized by MCS and EGO, whereas TF46 (Price with $N = 2$) could only be optimized by MCS, highlighting the impressive potential of MCS to optimize functions with complex profiles. TF22 and TF30 (Fig. 4e), TF25 (Fig. 4f for 2D surface plot for Rastrigin), and TF46 (Fig. 4g) have similar shape profiles with sinusoidal patterns embedded throughout a bowl-shaped basin. TF47 (Fig. 5h for 2D surface plot for Salomon) has a cone-shaped basin with ripple-like radial patterns. All these profiles are highly complex even for $N = 2$, thus their global optimization is hugely more challenging for $N > 2$.

Some multimodal TFs such as TF1 (Six Hump Camel Back with $N = 2$), TF43 (Exponential, $N = 2$), TF44 (Hosaki, $N = 2$), and TF45 (Miele Cantrell, $N = 4$) were relatively easy to optimize for all algorithms due to their smooth landscapes and well-separated optima.

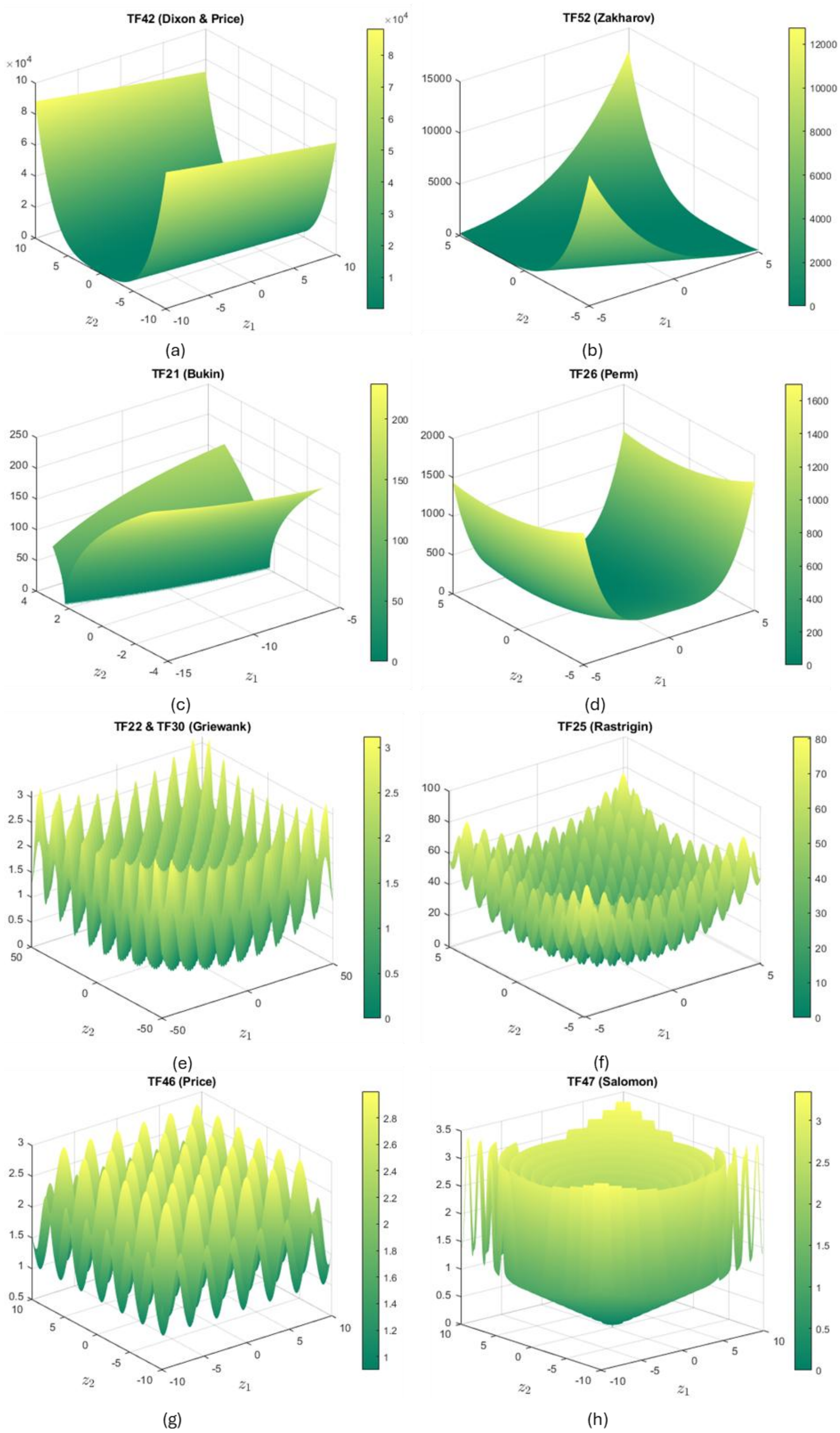

**Fig. 4:** 2D shape profiles of difficult test functions: (a) Dixon and Price, (b) Zakharov, (c) Bukin, (d) Perm, (e) Griewank, (f) Rastrigin, (g) Price, (h) Salomon

### 6.6 Comparison of Computational Run Times

While $\bar{\gamma}$ measures an algorithm's computational effort in terms of the number of function evaluations required to locate a global optimum, the time required to complete a single run of an algorithm provides an additional useful indicator of its applicability to implement for real-world applications. Barring IMPFIL and GLOBAL, which do not support termination based on a user-specified number of function evaluations, for all other algorithms, we record the wall-clock time ($\tau$) as the total time elapsed between the start and termination of an algorithm run after $K_{max}$ function evaluations for each TF. Table S5 in the supplementary material reports the median wall-clock time ($\tilde{\tau}$) over all runs for each test function and algorithm, excluding IMPFIL and GLOBAL. CGO and TSF are implemented only on TFs with $N \leq 4$ and $N \leq 6$, respectively; hence, their $\tilde{\tau}$ are not defined for higher-dimensional TFs.

Table 7 reports the average $\tilde{\tau}$ ($\bar{\tau}$) for each algorithm (except IMPFIL and GLOBAL) across all test functions for each $N$. The increasing trend of $\bar{\tau}$ with $N$ for all algorithms (except MCS and PSO) is expected, as the exponential growth of the search space with $N$ expands the area for domain exploration. For MCS, $\bar{\tau}$ marginally drops as $N$ increases from 3 to 4 and from 6 to 10, while for PSO, $\bar{\tau}$ drops as $N$ increases from 2 to 3 and from 6 to 10. However, the relatively small values of $\bar{\tau}$ for both MCS and PSO across $N$ makes these variations insignificant.

**Table 7:** $\bar{\tau}$ for each algorithm

| Algorithm | $N = 2$ | $N = 3$ | $N = 4$ | $N = 5$ | $N = 6$ | $N = 10$ |
|---|---|---|---|---|---|---|
| SBOC-R | 24.98 | 36.33 | 76.88 | 146.9 | 233.2 | 516.6 |
| SBOC-K | 37.32 | 63.05 | 117.9 | 170.1 | 332.5 | 1290 |
| GPS | 0.035 | 0.044 | 0.052 | 0.059 | 0.070 | 0.094 |
| MADS | 0.032 | 0.038 | 0.045 | 0.059 | 0.062 | 0.100 |
| DIRECT_G | 0.015 | 0.016 | 0.022 | 0.026 | 0.049 | 0.034 |
| MCS | 0.468 | 0.552 | 0.451 | 0.488 | 0.592 | 0.551 |
| GA | 0.344 | 0.397 | 0.506 | 0.520 | 0.521 | 0.726 |
| SA | 0.056 | 0.079 | 0.099 | 0.119 | 0.214 | 0.235 |
| PSO | 0.451 | 0.393 | 0.541 | 0.572 | 0.760 | 0.575 |
| BADS | 7.329 | 10.96 | 14.45 | 18.01 | 16.13 | 51.94 |
| SNOBFIT | 1.085 | 1.535 | 2.034 | 2.480 | 3.586 | 8.353 |
| CGO | 551.6 | 2596.5 | 3920 | - | - | - |
| MLMSRS | 17.13 | 26.82 | 37.96 | 55.5 | 60.9 | 140.2 |
| TSF | 42.31 | 63.26 | 183.5 | 1011 | 3438 | - |
| EGO | 24.22 | 79.41 | 161.3 | 602.4 | 1072 | 15355 |
| BAM | 169.0 | 238.7 | 332.1 | 413.3 | 514.2 | 871.7 |

**NOTE:** CGO and TSF cannot be run on TFs with $N > 4$ and $N > 6$, hence their $\bar{\tau}$ is undefined

Direct-search algorithms and metaheuristics algorithms are the fastest, completing each run within 1 second for all TFs. These algorithms rely on computationally cheap search and poll operations (as in direct search) and simple heuristic updates (as in metaheuristics), which entail modest computational overhead. On the contrary, SBO algorithms require significantly higher run times due to multiple time-intensive steps intrinsic to these algorithms, such as surrogate construction involving hyperparameter optimization, domain partitioning or exploration, and solving one or more sub-optimization problems.

Among the SBO algorithms, SNOBFIT required the least run time, completing a single run in less than 10 seconds for all TFs. BADS and MLMSRS were the next fastest, completing their runs within 1 minute and 2.5 minutes, respectively for all TFs. The small run time for SNOBFIT can be attributed to its use of simple local quadratic surrogates, whose construction and minimization is computationally inexpensive. BADS integrates a novel, fast, and efficient Bayesian optimization (BO) method within an MADS framework. Three key features of the BO step in BADS that contribute to its rapid execution are (1) re-training / updating the surrogate only when the predicted optimum does not improve, (2) optimizing the hyperparameters only when the model accuracy falls below a threshold, and (3) using a fast approximate optimization strategy to add new points. MLMSRS is based on computationally cheap RBF surrogates and employs a candidate sampling approach for adding new points without solving any explicit optimization sub-problems.

On the opposite end of the spectrum, CGO exhibited the highest run times for all TFs. $\bar{\tau}$ exceeded 9 minutes on two-dimensional TFs and increased exponentially on higher-dimensional TFs. A substantial increment in $\bar{\tau}$ was also observed for EGO on high-dimensional TFs ($N > 4$). Both algorithms solve one or multiple optimization sub-problems at each iteration, which substantially contributes to computational overhead. While CGO is based on RBF surrogates, EGO uses KRG surrogates which become increasingly expensive to train and optimize in higher dimensions.

While BAM exhibits relatively large wall-clock times for TFs across all $N$, it cannot be directly compared with other algorithms because BAM relies on an external executable for function evaluation, whereas all other algorithms call functions within the same runtime environment (MATLAB / Python). This introduces substantial overhead from repeated interpreter initialization and file-based input-output operations, which accumulate over iterations and inflates $\tau$ for a complete run. The observation that BAM's run time is already larger than other algorithms (except CGO) at small $N$ ($\leq 4$) and increases linearly (rather than exponentially) with increasing $N$ further suggests that the function evaluation overhead significantly influences $\bar{\tau}$.

SBOC-R, SBOC-K, and TSF require higher $\bar{\tau}$ than SNOBFIT, BADS, and MLMSRS, but lower than CGO, EGO, and BAM for all $N$ except $N = 10$, where SBOC-K required higher $\bar{\tau}$ than BAM. Both SBOC-R and SBOC-K required significantly lesser $\bar{\tau}$ than TSF. Between the two SBOC implementations, SBOC-K required higher run time than SBOC-R, since constructing KRG surrogates is computationally more expensive than RBF surrogates (Bhosekar and Ierapetritou, 2018).

To further analyze the time required by the key steps of SBOC, Table S6 in the supplementary file reports the median wall-clock times for surrogate training ($\tilde{\tau}_{ST}$), surrogate minimization ($\tilde{\tau}_{SM}$), cluster formation ($\tilde{\tau}_{CF}$), exploration ($\tilde{\tau}_{ER}$), and exploitation ($\tilde{\tau}_{EP}$) steps for both SBOC implementations and each TF. Table 8 lists the average wall-clock time ($\bar{\tau}_{ST}$, $\bar{\tau}_{SM}$, $\bar{\tau}_{CF}$, $\bar{\tau}_{ER}$, $\bar{\tau}_{EP}$) for these five steps across all TFs for each $N$. Our key findings from this breakdown of individual times for SBOC-R and SBOC-K are listed below.

1. The surrogate minimization step of SBOC consistently consumes the majority of the time for any run. It accounts for nearly 70% of the total wall-clock time for each $N$.

2. The exploration and exploitation steps require negligible time, each taking less than a second for all TFs. Between the two, exploitation is significantly faster than exploration.

3. The surrogate training step is more time-consuming for SBOC-K than for SBOC-R due to the larger computational overheads associated with constructing KRG surrogates.

4. The surrogate minimization step is more time-consuming for SBOC-K than SBOC-R. This can be attributed to the more complex analytical form for a KRG surrogate than an RBF surrogate, leading to increased optimization complexity.

5. As $N$ increases, the time required by each SBOC step increases. While this increase is marginal for exploration and exploitation, it becomes considerable for surrogate training, minimization, and clustering steps.

**Table 8:** Wall-clock times for individual steps of SBOC

| **SBOC-R** | | | | | |
|---|---|---|---|---|---|
| $N$ | **Surrogate Training** | **Surrogate Minimization** | **Cluster Formation** | **Exploration** | **Exploitation** |
| 2 | 2.175 | 15.88 | 6.650 | 0.177 | 0.012 |
| 3 | 4.061 | 23.96 | 7.951 | 0.251 | 0.012 |
| 4 | 9.438 | 52.49 | 14.28 | 0.496 | 0.024 |
| 5 | 23.11 | 96.97 | 25.71 | 0.856 | 0.039 |
| 6 | 37.05 | 163.6 | 31.15 | 1.049 | 0.050 |
| 10 | 128.9 | 333.5 | 51.85 | 1.777 | 0.086 |
| **SBOC-K** | | | | | |
| $N$ | **Surrogate Training** | **Surrogate Minimization** | **Cluster Formation** | **Exploration** | **Exploitation** |
| 2 | 2.180 | 27.44 | 6.908 | 0.454 | 0.020 |
| 3 | 5.613 | 48.57 | 10.14 | 0.667 | 0.027 |
| 4 | 16.52 | 86.29 | 15.98 | 0.775 | 0.030 |
| 5 | 28.21 | 103.0 | 18.31 | 0.701 | 0.025 |
| 6 | 61.31 | 206.6 | 35.81 | 1.175 | 0.051 |
| **SBOC-R** | | | | | |
| 10 | 163.1 | 997.5 | 70.63 | 2.460 | 0.145 |

Although our comparative analysis of algorithms based on $S$, $\overline{\Delta x}$, $\overline{\Delta f^*}$, $\bar{\gamma}$, and $\bar{\tau}$ offers a comprehensive assessment in terms of solution accuracy, computational effort, and run time, we additionally include information regarding the number of iterations required by an algorithm to determine an optimum with $\Delta f^* \leq \Delta f^{**}$ in the supplementary file. Table S7 reports the median number of iterations across all runs for all test functions and algorithms, except GLOBAL (which does not save its iterative history).

## 7. Conclusion

We presented a surrogate-based optimization algorithm (SBOC) for box-constrained systems based on the k-means clustering technique. Clusters were used to locate points in inter-cluster regions devoid of any sample points. SBOC is agnostic to the choice of surrogate modeling technique and terminates based on a user-specified limit on function evaluations. We used 52 diverse test functions from literature to assess its performance and compare it against promising benchmarking algorithms. We implemented SBOC using two surrogate forms, a Radial Basis Function form (SBOC-R) and a Kriging form (SBOC-K) and compared them with sixteen established optimization algorithms based on success rate, solution quality (proximity to nearest global optimum and deviation from the best function value) and computational effort. SBOC-K and SBOC-R were among the top two ranked algorithms based on success rates, successfully locating a global optimum for 65.4% and 63.5% of the test functions, respectively. Both consistently featured among the top six algorithms for solution quality and computational effort. They not only estimated the global optimum accurately for most test functions but also converged close to a global optimum for most. SBOC-K ranked second while SBOC-R ranked fourth among all algorithms based on computational effort. They performed especially well on test functions with four or more input variables. They outperformed all others in optimizing these difficult functions to global optimality.

SBOC stands out as a promising, generic, robust, and computationally efficient surrogate-based optimization algorithm for global optimization of box-constrained systems. However, it is difficult to get a perfect algorithm. Shapes and profiles of test functions obviously affect the ability of global optimization algorithms. Some multimodal test functions with highly complex, nonlinear profiles posed much difficulty for all algorithms and could not be optimized successfully by any algorithm within the specified computational budget. In contrast, some others with smoother and simpler shape profiles were optimized by all algorithms. Our extensive analysis of computational times revealed that surrogate-based optimization algorithms generally require longer run times than direct search and metaheuristics approaches. Among the surrogate-based optimization algorithms considered in our analysis, SBOC demonstrated moderate run time requirements across test functions of varying dimensionality. The choice of the surrogate form naturally affects run times. Since Kriging surrogates are computationally more expensive to build than Radial Basis Function models, SBOC-K required larger run times than SBOC-R.

A natural extension of this work is to modify SBOC to tackle large-scale, complex constrained problems with continuous and discrete variables. While the core framework may remain largely unchanged, this adaptation would require deliberate consideration of key challenges and issues such as surrogate modeling for mixed-variable systems, efficient handling of constraints, point addition for continuous and discrete variables, and computational efficiency of the revised algorithm. This extension holds strong potential to further enhance the scope and applicability of SBOC to real-world systems.

## Acknowledgement

This study was supported by the Agency for Science, Technology and Research, Singapore under its Low-Carbon Energy Research Funding Initiative (U2102d2003).

## Supplementary Information

Supplementary material is associated with the published journal article at https://doi.org/10.1016/j.compchemeng.2026.109559 (see the Supplementary material linked from the article page).